\documentclass[preprint,review,10pt]{elsarticle}
\usepackage{amsmath}
\usepackage{lineno}
\usepackage{graphicx}    
\usepackage{algorithm}
\usepackage{algpseudocode}
\usepackage{subfigure} 
\usepackage{setspace}  
\usepackage{color}
\usepackage{graphicx}
\usepackage{amssymb}
\usepackage{array}
\usepackage{subfigure}
\usepackage{booktabs}
\usepackage{multirow}
\usepackage{color}
\usepackage{bigstrut}
\usepackage{caption}
\usepackage{threeparttable}
\usepackage{bm}
\newtheorem{remark}{Remark}
\newtheorem{example}{Example}
\usepackage{color,xcolor}

\newproof{pf}{Proof}

\journal{Knowledge-Based Systems}

\begin{document}

\begin{frontmatter}



\title{TECM: Transfer Learning-based Evidential C-Means Clustering}


\author{Lianmeng Jiao$^\ast$}
\ead{jiaolianmeng@nwpu.edu.cn}
\author{Feng Wang}
\ead{fengwang@mail.nwpu.edu.cn}
\author{Zhun-ga Liu}
\ead{liuzhunga@nwpu.edu.cn}
\author{Quan Pan}
\ead{quanpan@nwpu.edu.cn}

\cortext[cor1]{Corresponding author at: School of Automation, Northwestern Polytechnical
University, Xi'an 710072, P.R. China. Tel.: +86 29 88431306; Fax: +86 29 88431306.}

\address{School of Automation, Northwestern Polytechnical University, Xi'an 710072, P.R. China}

\begin{abstract}
As a representative evidential clustering algorithm, evidential c-means (ECM) provides a deeper insight into the data by allowing an object to belong not only to a single class, but also to any subset of a collection of classes, which generalizes the hard, fuzzy, possibilistic, and rough partitions. However, compared with other partition-based algorithms, ECM must estimate numerous additional parameters, and thus insufficient or contaminated data will have a greater influence on its clustering performance. To solve this problem, in this study, a transfer learning-based ECM  (TECM) algorithm is proposed by introducing the strategy of transfer learning into the process of evidential clustering. The TECM objective function is constructed by integrating the knowledge learned from the source domain with the data in the target domain to cluster the target data. Subsequently, an alternate optimization scheme is developed to solve the constraint objective function of the TECM algorithm. The proposed TECM algorithm  is applicable to cases where  the source and target domains have the same or different numbers of clusters. A series of experiments were conducted on both synthetic and real datasets, and the experimental results demonstrated the effectiveness of the proposed TECM algorithm  compared to ECM and other representative multitask or  transfer-clustering algorithms.
\end{abstract}


\end{frontmatter}

\section{Introduction}
%
%
%
%

Clustering is a process of dividing a set of objects into groups of similar members. Classical clustering algorithms can be roughly divided into the following categories: partition-based methods \cite{chang2017,unified2021}, hierarchy-based methods \cite{Hierarchical2020,Parallel2019}, density-based methods \cite{Multi-Density2019,ant2019}, grid-based methods  \cite{Online2020,grid2018}, and graph-based methods \cite{GMC2020,Graph2018}. Among these, partition-based methods are the most popular because they are universally applicable to numerous real-world clustering tasks. Although classical partition-based clustering algorithms such as c-means only provide hard partitions, a large number of variants have been proposed by incorporating fuzzy sets \cite{Kernel2016}, possibility theory \cite{fully-unsupervised2018}, and rough sets \cite{rough k-means} to characterize the uncertainty of cluster membership.
Recently, evidential clustering \cite{2008ECM,2015ECM,ECM2016,Evidential2016,2020EGMM,NN-EVCLUS2021,Dynamic2021} has been developed as a promising framework that uses the Dempster-Shafer belief function theory \cite{Upper1967,Shafer1976} as an uncertainty model. It generalizes the above-mentioned hard, fuzzy, possibilistic, and rough partitions, and thus, has been widely used in numerous application scenarios, including  tumor segmentation \cite{Spatial2017,Joint2018}, community detection \cite{Median2015,SELP2018}, and items of interest recommendation \cite{evidential2018,evidential2019}.

Evidential c-means (ECM) \cite{2008ECM}, as a representative evidential clustering algorithm, can assign objects not only to a single class,  but also to any subset of a collection of classes by credal partition \cite{2004EVCLUS}. Therefore, ECM can provide a deeper insight into the data for situations where the data distribution is complex.
However, this additional flexibility comes at the expense of algorithm complexity, that is, the ECM algorithm must estimate  numerous additional parameters compared to  the original c-means or fuzzy c-means (FCM) algorithms; therefore, insufficient or contaminated data will have a greater influence on its clustering performance. Noise is unavoidable in practical clustering. In addition, collecting sufficient data is difficult in certain practical applications. Such shortcomings severely restrict the applicability of ECM. One feasible method to address this problem is transfer learning \cite{survey2015,survey2021}.

Transfer learning assumes that the data in the current scene (called the target domain) are inadequate for the learning task, whereas other helpful information is available from a relevant scene (called the source domain). Currently, transfer learning is mainly integrated with the following machine learning tasks \cite{pan2010survey}: classification \cite{minimum2012,Adaptive2008}, regression \cite{Knowledge2013,Bayesian2008}, feature
extraction \cite{Domain2011,Transferred2008}, and clustering \cite{qian2015cluster,deng2015transfer}. The first three tasks have been studied extensively; however, research on integrating transfer learning with clustering remains inadequate, although it has wide application in reality.
 In the past decade, research on transfer clustering has gradually emerged. Studies on transfer clustering can be broadly classified into the following four categories based on the transfer strategy \cite{Survey2021}: (1) the instance-based method \cite{2013Transfer}, where  a portion of the source data is selected and transferred to the target domain for clustering target data; (2) the feature-representation-based method \cite{yang2017common,dai2008self}, whose intuitive idea is to generate a superior feature representation for the  target data based on the knowledge learned from the source data; (3) the parameter-based method \cite{qian2015cluster,deng2015transfer,gaussian2022,qian2017knowledge,cheng2017maximum,gargees2020tlpcm}, which assumes that two domains share similar parameters such as cluster centers; and (4) the relational knowledge-based method \cite{spectral2012,yu2012transfer}, which builds a  mapping of the relational knowledge between two domains. Among these, parameter-based transfer clustering is the focus of the current research because it can fully exploit the  inter-domain similarity and is more suitable for integration with popular partition-based clustering algorithms.

In this paper\footnote{A preliminary version of this paper have been presented in BELIEF 2021 Conference \cite{Jiao21}. This paper extends our short conference paper by adding a comprehensive review of related work, a deep analysis of the challenges, a more detailed description of the proposed algorithm with an analysis of its generality, convergence, and complexity, and additional  datasets and comparison algorithms to illustrate the advantages of the proposed algorithm.}, transfer learning is applied to ECM to improve the clustering performance when the target data are insufficient or contaminated. A  parameter-based transfer strategy is adopted to develop a transfer learning-based ECM (TECM) algorithm, that is, the parameters induced from the source data, namely, barycenters, are appropriately used to improve the clustering performance of the target data. The main contributions of this study are summarized as follows.

(1) A novel TECM objective function is constructed by introducing the barycenters learned from the source data as knowledge to guide the clustering of the target data.

(2) An alternate optimization scheme is developed to solve the constrained TECM objective function, which derives closed-form solutions by alternately updating the parameters.

(3) The proposed TECM is applicable to any transfer-clustering scenario where the source and target domains have the same or different numbers of clusters.

A series of experiments on both synthetic and real datasets are  designed to demonstrate the effectiveness of the TECM algorithm  proposed in this study. In the experiments on synthetic datasets, considering three situations (the number of clusters in the source domain is equal to, greater than, or less than the number of clusters in the target domain) and two application scenarios (target data are insufficient or contaminated), six groups of synthetic datasets are designed to illustrate the TECM  clustering performance. Furthermore, two synthetic ambiguous datasets are designed to demonstrate the superiority of TECM in characterizing the uncertainty of cluster membership.
In the experiments on real datasets, the segmentation results of texture images are used to illustrate the TECM clustering performance compared to ECM and other representative multitask or transfer-clustering algorithms.
The experimental results on the synthetic and real datasets demonstrate that the proposed TECM achieves superior clustering performance compared to  the other methods on the three popular validity indices.

The remainder of this paper is organized as follows. Section \ref{two} comprehensively reviews of the related work. The ECM algorithm is summarized and analyzed in Section \ref{three}. A detailed description of the proposed TECM is presented in Section \ref{four}. In Section \ref{five}, we evaluate the proposed TECM using both synthetic and real datasets. Section \ref{six} concludes the paper and outlines the  directions for future research.

\section{Related work} \label{two}
In the past decade, transfer-learning techniques have been integrated with classical clustering algorithms. Based on their transfer strategy, the existing transfer-clustering  algorithms can be categorized as follows.

\textit{Instance-based method}. In \cite{2013Transfer}, the unlabeled heterogeneous data from different domains are first projected into a common subspace, and then other informative source data are selected and transferred to the target domain to improve the clustering performance of the target data using an algorithm called DicTrans. However, because of the differences in data distribution between the two domains, it is difficult to determine how to select the source data and the amount of source data to be transferred.

\textit{Feature-representation-based method}. In \cite{yang2017common}, based on co-occurrence data, a co-transfer-clustering algorithm was presented to simultaneously address multiple individual clustering tasks. A collective nonnegative matrix factorization algorithm is used to obtain a common subspace for all domains. Then, data in different domains are mapped into this subspace and the data from the different domains are clustered simultaneously using a symmetric nonnegative matrix factorization algorithm.
In \cite{dai2008self}, a transfer-clustering algorithm called shared hidden space transfer fuzzy c-means was proposed. In this algorithm, source data and target data are mapped into a common subspace by attempting to identify a projected matrix with the strategy of maximum mean discrepancy; then, all data in the common subspace are clustered simultaneously.

\textit{Parameter-based method}. In \cite{deng2015transfer}, a representative parameter-based transfer-clustering algorithm called transfer fuzzy c-means (TFCM) clustering was proposed using the transfer-learning strategy. The cluster centers obtained from the source domain via classical FCM clustering are used to construct the corresponding objective function to promote the clustering performance of the target domain.
In \cite{qian2017knowledge,cheng2017maximum,gargees2020tlpcm}, as  in \cite{deng2015transfer}, the cluster centers obtained from the source domain are used as transfer knowledge to guide the clustering of the target domain.
In \cite{gaussian2022}, a novel transfer-clustering algorithm called transfer distributed EM clustering (TDEM) was proposed, where each node is considered as both the source and target domains;  these nodes can learn from each other to complete the clustering task in distributed peer-to-peer (P2P) networks.
In \cite{qian2015cluster}, a transfer maximum entropy clustering algorithm was proposed and the corresponding objective function was constructed using the cluster centers and membership matrix obtained from the source data. However, this algorithm is only applicable to situations where the source and target domains have the same number of clusters.

\textit{Relational knowledge-based method}. In \cite{spectral2012}, transfer spectral clustering (TSC), a representative relational knowledge-based transfer-clustering algorithm, was proposed. Based on the assumption that embedded feature information can connect two clustering tasks, TSC can simultaneously improve the clustering performance in two domains. Furthermore, co-clustering is employed to control the knowledge transfer between the two tasks. In \cite{yu2012transfer}, a novel transfer-clustering algorithm called topic constraint transfer clustering was proposed. First, several clusters are obtained from the source data using an efficient clustering algorithm. Subsequently, several topics are extracted from each cluster. Then, the similarities between the target data and extracted topics are discovered to assist in the clustering of the target data. Finally, a semi-supervised clustering algorithm is applied to the target data.

Apart from the above four categories of transfer-clustering  algorithms, our task is also related to multitask clustering, which addresses multiple clustering tasks simultaneously and uses mutual cooperation to improve the performance of all the  clustering tasks. Representative multitask clustering algorithms include learning the shared subspace for multi-task
clustering (LSSMTC)  and combining k-means (CombKM)  \cite{2009Learning}. LSSMTC aims to learn a subspace shared by all tasks, through which the knowledge of the tasks can be transferred to each other;  CombKM clusters all data from different tasks simultaneously using c-means.

Among these transfer or multitask clustering methods, parameter-based transfer clustering is the focus of the current research because it is more suitable for integration with popular partition-based clustering algorithms. In partition-based clustering algorithms, the cluster membership of each object depends on its distance to the cluster centers; therefore, the parameter ``cluster centers''  is the most reliable and intuitive transfer knowledge. In addition, parameter-based transfer clustering typically obtains acceptable transfer performance because it can fully utilize the  inter-domain similarity. Therefore, the parameter-based transfer strategy is  adopted in this study to develop an ECM transfer-clustering   version.

\section{ECM: Basics and challenges} \label{three}
Belief function theory \cite{Upper1967,Shafer1976} is a formal framework for approximate reasoning.
In this theory, the set $\Omega$ = $\{w_1 ,...,w_c\}$ containing all possible results of a discrimination problem is called the frame of discernment.
A mass function $m$ is used to express uncertain information, which is a mapping function from $2^\Omega$ to [0, 1] that satisfies the following equations:
\begin{equation}
m(\emptyset ) = 0,\;\sum\limits_{A \in {2^\Omega }} {m(A) = 1}.
\label{eq1}
\end{equation}
Subsets $A$ are called focal sets if $A$ $\in$ $2^\Omega$ satisfies $m(A) > 0$. Based on the type of information represented, the mass function $m$ is classified into the following categories:

$\bullet$ \emph{Bayesian}: The focal sets of this type of mass function are singletons. In this case, the mass function becomes the classical probability distribution function.

$\bullet$ \emph{Certain}: This type of mass function has only one focal set, which can be denoted by $m_{A}$, where $A$ is its unique focal set. This mass function certainly indicates that the real result is in set $A$.

$\bullet$ \emph{Vacuous}: This type of mass function has only one focal set, $\Omega$, which can be denoted as $m_{\Omega}$. This mass function indicates that the actual result is unknown.

In \cite{2004EVCLUS}, the notion of a credal partition was proposed to represent the class membership uncertainty of each object $x_i$ by a mass function $m_i$, which can model different situations from completely unknown to total certainty regarding the class membership of the object.
\begin{example}
	 Consider an example of $n = 5$ objects belonging to $c = 2$ classes with evidential membership. The mass functions of these objects are listed in Table~\ref{tab0}, which illustrates all possible situations from complete unknown to total certainty.
	 The case of object $1$ corresponds to the situation of total certainty ($m_1$ is certain).
	 Object $2$ corresponds to the situation of completely unknown ($m_2$ is vacuous).
	 Object $3$ corresponds to the situation of probabilistic uncertainty ($m_3$ is Bayesian).
	 Object $4$ is a general example of evidential membership.
	 Mass $m_5(\emptyset)=1$ indicates that the class of object $5$ is not in $\Omega$.
\end{example}

\begin{table}[htbp]
	\centering
	\caption{Illustration of credal partition}
	\begin{tabular}{|c|c|c|c|c|c|}
		\hline
		$A$            & $m_1(A)$        & $m_2(A)$   &$m_3(A)$     & $m_4(A)$        & $m_5(A)$ \bigstrut\\
		\hline
		$\emptyset$            & 0            & 0      &0      & 0.1            & 1 \bigstrut[t]\\
		\hline	{$w_1$}         & 1            & 0      & 0.4     & 0.4           & 0 \bigstrut[t] \\
		\hline	{$w_2$}         & 0            & 0      & 0.6     & 0.3            & 0 \bigstrut[t] \\
		\hline	{$\Omega$}   & 0            & 1     &  0     & 0.2          & 0 \bigstrut[t]\\
		\hline
	\end{tabular}%
	\label{tab0}%
\end{table}%

The ECM algorithm \cite{2008ECM} is a representative clustering algorithm with credal partition, which is an extended version of FCM in the framework of belief function theory. The mass ${m_{ij}}={m_i}({A_j})$ associated with object $x_i$ with $A_j$ $({A_j} \ne \emptyset ,{\kern 1pt} \,{A_j} \subseteq \Omega)$ depends on the distance ${d_{ij}} = ||{x_i} - {\bar v_j}|{|^2}$ between object $x_i$ and barycenter ${\overline v _j}$ of $A_j$, which is greater when $d_{ij}$ is smaller. Barycenter  ${\overline v _j}$ is defined as
\begin{equation}
{\overline v _j} = \frac{1}{{{c_j}}}\sum\limits_{k = 1}^c {{s_{kj}}{v_k}},\quad {\rm{s}}.{\kern 1pt} \,{\rm{t}}.\quad s_{kj} = \left\{ \begin{array}{l}
1\quad {\rm{if}}\;{w_k} \in {A_j}\\
0\quad \rm{else},
\end{array} \right.
\label{eq3}
\end{equation}
where $c_j$ is the cardinality of the focal set $A_j$ and $v_k$ is the center of the singleton $w_k$.

For $n$ objects with $p$ dimensions, the ECM objective function with respect to the credal partition $M = \{ {m_1},...,{m_n}\}  \in {R^{n \times {2^c}}}$ and cluster centers $V=\{ {v_1},...,{v_c}\} \in R^{c \times p}$ is defined as
\begin{equation}
\begin{split}
&\mathop {\min }{J_{ECM}(M,V)} = \sum\limits_{i = 1}^n {\sum\limits_{\{ j/{A_j} \subseteq \Omega, {A_j} \ne \emptyset \} } {c_j^\alpha m_{ij}^\beta ||{x_i} - {{\overline v }_j}|{|^2}} } + \sum\limits_{i = 1}^n {{\delta ^2}m_{i\emptyset }^\beta},
\\&{\rm{s}}.{\kern 1pt} \,{\rm{t}}.\quad \sum\limits_{\{ j/{A_j} \subseteq \Omega ,{A_j} \ne \emptyset \} } {{m_{ij}} + {m_{i\emptyset }} = 1, \quad \forall i = 1,...,n,}
\end{split}
\label{eq5}
\end{equation}
where ${m_{i\emptyset }}$ represents ${m_i}(\emptyset )$ and $c_j$ represents the cardinality of the focal set $A_j$. Parameters $\alpha$ and $\beta$ control the penalization degree of high cardinality and credibility of the partition, respectively, and $\delta$ controls the number of outliers.

Based on the objective function given in Eq. ~(\ref{eq5}), it can be observed that the ECM must estimate a considerably greater number of parameters ($n * 2^c$) to obtain a credal partition compared to other classical partition-based algorithms ($n * c$ for FCM and $n$ for c-means). Therefore, insufficient or contaminated data could have a greater influence on ECM compared to other partition-based algorithms. Here, we provide an example to illustrate this issue.

\begin{figure}[!htb]
	\centering
	\includegraphics[width=\hsize]{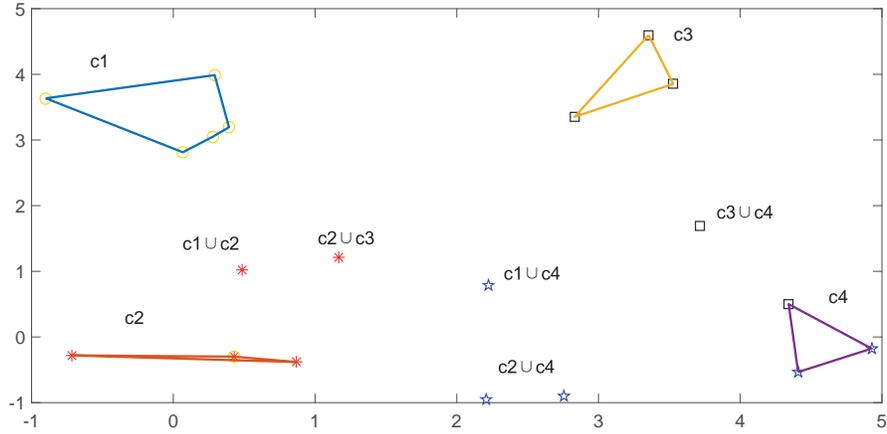}
	\caption{Clustering results for dataset $Synt_1$ using ECM.}
	\label{fig1}
\end{figure}

\begin{figure}[!htb]
	\centering
	\includegraphics[width=\hsize]{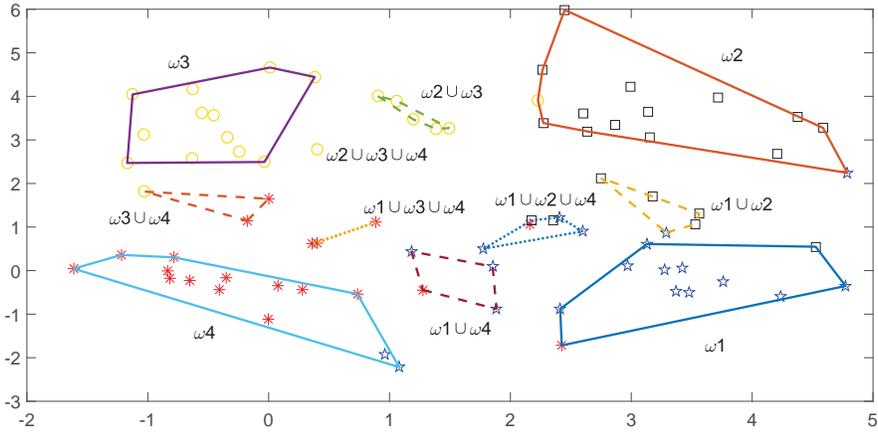}
	\caption{Clustering results for dataset $Synt_2$ using ECM.}
	\label{fig2}
\end{figure}

\begin{example}
	Two synthetic datasets, $Synt_1$ and $Synt_2$, are used to illustrate the ECM clustering performance for application scenarios where the data are insufficient and contaminated, respectively. Let \bm{$\mu_{i}$} and \bm{$\Sigma_{i}$} represent the mean vector and covariance matrix of the $i$th cluster in each dataset, respectively. Dataset $Synt_1$ simulates a scenario where the data are insufficient. Dataset $Synt_1$ is generated with \bm{$\mu_{1}}=[0, 0]$, \bm{$\Sigma_{1}}=[1, 0;0, 1]$,  \bm{$\mu_{2}}=[0, 3]$, \bm{$\Sigma_{2}}=[1, 0;0, 1]$, \bm{$\mu_{3}}=[3, 0]$, \bm{$\Sigma_{3}}=[1, 0;0, 1]$, \bm{$\mu_{4}}=[3, 3]$, and  \bm{$\Sigma_{4}}=[1, 0;0, 1]$, where  each cluster consists of only  five objects. The data distributions used to generate the dataset $Synt_2$ are the same as those for $Synt_1$; however, each cluster in $Synt_2$ consists of 20 objects. In addition, zero-mean Gaussian noise with a standard deviation of 0.3 is added to $Synt_2$ to simulate the scenario where the data are contaminated. The clustering results for datasets $Synt_1$ and $Synt_2$ using ECM are displayed  in Fig. ~\ref{fig1} and Fig. ~\ref{fig2}, respectively. It can be observed from the clustering results that the real cluster profiles were not recovered precisely, and the ambiguous objects at the boundaries were not properly recognized.
\end{example}

The above example confirms that the ECM  clustering performance is not ideal for situations where the data are insufficient or contaminated. Therefore, it is necessary to improve the ECM  clustering performance in these situations by exploiting sufficient high-quality data from a relevant scene.

\section{TECM} \label{four}
In this section, a TECM clustering algorithm is developed to improve the clustering performance when the target data are insufficient or contaminated. In Section \ref{4.1}, the basic principle of TECM is introduced. In Section \ref{4.2}, a novel TECM objective function is presented. In Section \ref{4.3}, a procedure for solving the TECM  objective function is developed. In Section \ref{4.4}, we provide the algorithm pseudocode and analyze the generality, convergence, and complexity of the proposed algorithm.

\subsection{Basic principle of TECM} \label{4.1}
\begin{figure}[!htb]
	\centering
	\includegraphics[width=0.9\hsize]{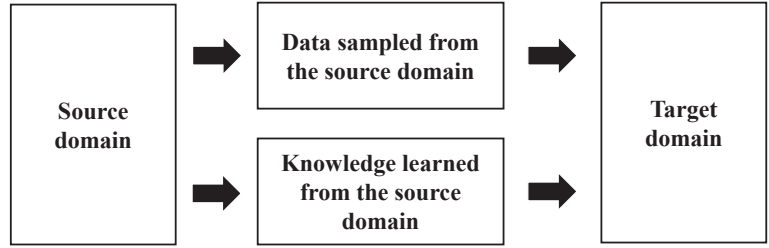}
	\caption{Available information obtained from source data for clustering of target data.}
	\label{fig3}
\end{figure}

In transfer learning, as indicated in Fig. ~\ref{fig3}, sampled data and learned knowledge are typically the two types of information available from the source data.
The sampled data are not appropriate for the direct clustering of the target data because of the differences in the data distribution between these two domains. In addition, source data are at times inaccessible because of privacy protection in certain practical scenarios.
Compared with sampled data, learned knowledge such as feature representations, parameters, and relationships is typically considered more insightful and noise-resistant.

\begin{figure}[!htb]
	\centering
	\includegraphics[width=0.9\hsize]{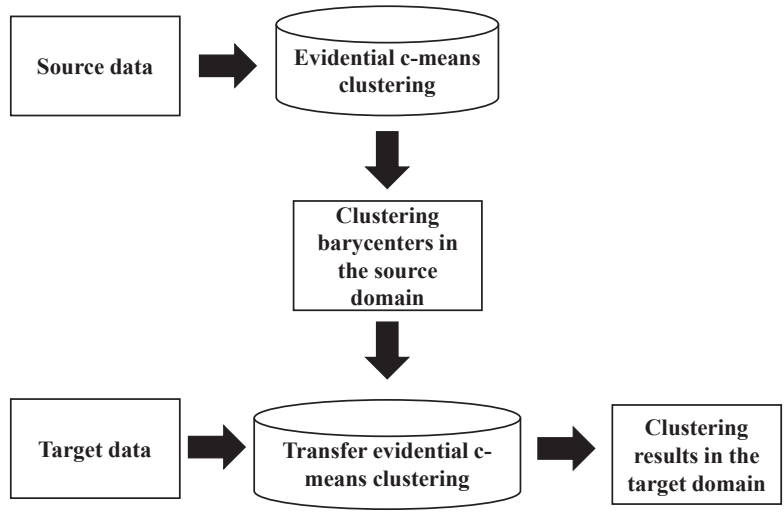}
	\caption{Basic principle of TECM.}
	\label{fig4}
\end{figure}

In this study, the TECM clustering algorithm was developed using the learned knowledge from sufficient data of acceptable quality in a relevant scene. The basic principle of TECM is illustrated in Fig. ~\ref{fig4}. It can be observed that ECM is first applied to the source data to learn the knowledge, that is, barycenters of the clusters, which concisely represent the data distribution. These barycenters are then used to guide the clustering of the target data. Following this principle, we study how to construct the TECM objective function by appropriately using the knowledge learned from the source data and how to develop an efficient optimization scheme to solve the objective function.

\subsection{TECM Objective function} \label{4.2}
In this section, the TECM objective function is constructed by introducing the learned barycenters into the ECM objective function. The TECM objective function with regard to the credal partition $M$, association matrix $R$, and cluster centers $V$ is defined as

\begin{equation}
\begin{split}
&\mathop {\min }{J_{TECM}(M,R,V)}= \sum\limits_{i = 1}^n {\sum\limits_{\{ j/{A_j} \subseteq \Omega_t, {A_j} \ne \emptyset\} } {c_j^\alpha m_{ij}^\beta ||{x_i} - {{\bar v}_j}|{|^2}} }  \hfill
+\\& \lambda \sum\limits_{\{ k/{A_k} \subseteq \Omega_s, {A_k} \ne \emptyset\} } {\sum\limits_{\{ j/{A_j} \subseteq \Omega_t, {A_j} \ne \emptyset\} } {c_j^\alpha r_{kj}^\gamma ||{{\tilde v}_k} - {{\bar v}_j}|{|^2}} }  \hfill
+ \sum\limits_{i = 1}^n {{\delta ^2}m_{i\emptyset }^\beta }  \hfill,
\\&s.t.{\kern 1pt} \;\sum\limits_{\{ j/{A_j} \subseteq \Omega_t, {A_j} \ne \emptyset\} } {{m_{ij}} + {m_{i\emptyset }} = 1,\;\;\forall i = 1,...,n,} \\&\hfill  {\kern 1pt} \;\sum\limits_{\{ j/{A_j} \subseteq {\Omega _t},{A_j} \ne \emptyset\} } {{r_{kj}} = 1,\quad \forall k/} {A_k} \subseteq {\Omega _s},{A_k} \ne \emptyset,\\
\end{split}
\label{eq6}
\end{equation}
where ${\widetilde v_k}$ is the $k$-th barycenter obtained from the source data and element $r_{kj}$ of $R$ denotes the association between the $j$-th barycenter learned from the target data and $k$-th barycenter learned from the source data. The sets $\Omega_t$ and $\Omega_s$ denote the frames of the discernment of the target and source domains, respectively. Parameter $\gamma$ controls the credibility of the association and $\lambda$ is the balance coefficient of transfer learning. The definitions of the other parameters $\alpha$, $\beta$, and $\delta$ are the same as those in ECM.

The characteristics of the objective function defined in Eq. ~(\ref{eq6}) are as follows.

1) The first and third terms are inherited from the ECM  objective function, which is used to generate the credal partition of the target data by minimizing the global inner-cluster distance by considering the membership degrees between the objects and barycenters.

2) The second term introduces knowledge learned from the source data. The barycenters generated from the source data provide concise yet  rich class distribution information of the relevant source domain. This term requires that the barycenters of the target data be as close as possible to the barycenters learned from the source data as a whole by considering their potential associations.

3) The parameter $\lambda$ is a balance coefficient used to control the transfer degree of the learned knowledge from the source data. If the value of $\lambda$ is zero, TECM becomes the original ECM; if the value of $\lambda$ is relatively large, a negative transfer could  occur. Therefore, an appropriate value of $\lambda$ must be selected. The choice of $\lambda$ depends on the specific application; the grid search strategy \cite{Deng2010} can be used to determine an appropriate value. A series of experiments \cite{deng2015transfer}  proved that as the value of $\lambda$ increases, the clustering performance always improves first and then worsens. Based on this, we can first set a large search range with a large step size, and then select the part with superior  clustering performance and reduce the step size. These steps are repeated to obtain an appropriate value of $\lambda$ with acceptable clustering performance.
Parameter $\gamma$ can be set to $2$, similar to that in \cite{deng2015transfer}. The values of the other parameters $\alpha$, $\beta$, and $\delta$ can be set to the same values as in ECM \cite{2008ECM}.

4) The objective function is defined for the general case, where  the frames of discernment $\Omega_s$ and $\Omega_t$ can be different. By minimizing the objective function, we can determine the optimal pairwise associations of the barycenters (represented by the association matrix $R$) between the source and target data. This property significantly  expands the scope of application of TECM.

\subsection{Optimization procedure} \label{4.3}
To minimize $J_{TECM}$, an alternate optimization scheme is used in this study, that is, optimizing one variable by setting the other variables to fixed values. From Eq.~(\ref{eq6}), it can be observed that the objective function $J_{TECM}$ is unconstrained to $V$, yet is constrained to $M$ and $R$. To solve the constrained object function for $M$ and $R$, $n$ Lagrange multipliers $\lambda_i$ and $k$ ($k$ is the number of focal sets in $\Omega_s$) Lagrange multipliers $\eta _k$ are introduced to formulate the following Lagrangian objective function $J_L$:

\begin{equation}
\begin{split}
{J_L} = {J_{TECM}}& + \sum\limits_{i = 1}^n {\lambda _i}\left[ {1 - (\sum\limits_{\{ j/{A_j} \subseteq \Omega_t ,{A_{j \ne \emptyset }}\} } {{m_{ij}} + {m_{i\emptyset }}} )} \right]\\& + \sum\limits_{\{k/{A_k} \subseteq \Omega_s, {A_k} \ne \emptyset\}}\ {{\eta _k}} (1 - \sum\limits_{\{ j/{A_j} \subseteq \Omega_t ,{A_{j \ne \emptyset }}\} } {{r_{kj}}} ).
\end{split}
\label{eq7}
\end{equation}

\subsubsection{Optimization with respect to credal partition $M$}
 First, let $V$ and $R$ be fixed. Taking the partial derivatives of $J_L$ to $m_{ij}$, $m_{i\emptyset}$, and $\lambda_i$, and setting them to $0$, we obtain

\begin{equation}
\begin{split}
\frac{{\partial J_L}}{{\partial {m_{ij}}}} = \beta c_j^\alpha m_{ij}^{\beta  - 1}d_{ij}^2 - {\lambda _i} = 0,
\end{split}
\label{eq8}
\end{equation}

\begin{equation}
\begin{split}
\frac{{\partial J_L}}{{\partial {m_{i\emptyset }}}} = \beta {\delta ^2}m_{i\emptyset }^{\beta  - 1} - {\lambda _i} = 0,
\end{split}
\label{eq9}
\end{equation}

\begin{equation}
\begin{split}
\frac{{\partial J_L}}{{\partial {\lambda _i}}} = \sum\limits_{\{ j/{A_j} \subseteq \Omega_t ,{A_j} \ne \emptyset \} } {{m_{ij}} + {m_{i\emptyset }} - 1}  = 0.
\end{split}
\label{eq10}
\end{equation}

From Eq.~(\ref{eq8}), we obtain
\begin{equation}
\begin{split}
{m_{ij}} = {\left( {\frac{{{\lambda _i}}}{\beta }} \right)^{\frac{1}{{\beta  - 1}}}}{\left( {\frac{1}{{c_j^\alpha ||{x_i} - {{\bar v}_j}|{|^2}}}} \right)^{\frac{1}{{\beta  - 1}}}},
\end{split}
\label{eq11}
\end{equation}
and from Eq.~(\ref{eq9}), we obtain
\begin{equation}
{m_{i\emptyset }} = {\left( {\frac{{{\lambda _i}}}{\beta }} \right)^{\frac{1}{{\beta  - 1}}}}{\left( {\frac{1}{{{\delta ^2}}}} \right)^{\frac{1}{{\beta  - 1}}}}.
\label{eq12}
\end{equation}

Using Eq.~(\ref{eq10})--Eq.~(\ref{eq12}), we obtain
\begin{equation}
\begin{split}
{\left( {\frac{{{\lambda _i}}}{\beta }} \right)^{\frac{1}{{\beta  - 1}}}} = {\left( {\sum\limits_j {\frac{1}{{{{(c_j^\alpha ||{x_i} - {{\bar v}_j}|{|^2})}^{\frac{1}{{\beta  - 1}}}}}} + \frac{1}{{{\delta ^{\frac{2}{{\beta  - 1}}}}}}} } \right)^{ - 1}}.
\end{split}
\label{eq13}
\end{equation}

Substituting Eq.~(\ref{eq13}) into Eq.~(\ref{eq11}), we obtain the updated equation for $M$:

\begin{equation}
\begin{split}
&{m_{ij}} = \frac{{{{\left( {c_j^\alpha {\rm{||}}{x_i} - {{\bar v}_j}|{|^2}} \right)}^{\frac{{ - 1}}{{\beta  - 1}}}}}}{{\sum\nolimits_{{A_t} \ne \emptyset } {{{\left( {c_t^\alpha ||{x_i} - {{\bar v}_t}|{|^2}} \right)}^{\frac{{ - 1}}{{\beta  - 1}}}} + {\delta ^{\frac{{ - 2}}{{\beta  - 1}}}}} }},\\
&\forall i = 1,...,n,\quad \forall j/{A_j} \subseteq {\Omega _t},{A_j} \ne \emptyset ,
\end{split}
\label{eq14}
\end{equation}
and
\begin{equation}
\begin{split}
{m_{i\emptyset }} =& 1 - \sum\limits_{j/{A_j} \subseteq {\Omega _t},{A_j} \ne \emptyset } {{m_{ij}}},~~~~ \forall i = 1,...,n.
\end{split}
\label{eq15}
\end{equation}

\subsubsection{Optimization with respect to association matrix $R$}
Next, we assume that $V$ and $M$ are fixed. Taking the partial derivatives of $J_L$ to $r_{kj}$ and $\eta _k$ and setting them to $0$, we obtain

\begin{equation}
\frac{{\partial J_L}}{{\partial r{}_{kj}}} = \lambda c_j^\alpha \gamma r_{kj}^{\gamma  - 1}{({{\tilde v}_k} - {v_j})^2} - {\eta _k} = 0,
\label{eq16}
\end{equation}

\begin{equation}
\frac{{\partial J_L}}{{\partial {\eta _k}}} = \sum\limits_{\{ j/{A_j} \subseteq \Omega_t ,{A_j} \ne \emptyset \} } {{r_{kj}} - 1 = 0.}
\label{eq17}
\end{equation}

From Eq.~(\ref{eq16}), we obtain

\begin{equation}
{r_{kj}} = {\left( {\frac{{{\eta _k}}}{{{\lambda _\gamma }}}} \right)^{\frac{1}{{\gamma  - 1}}}}{\left( {\frac{1}{{c_j^\alpha ||{{\tilde v}_k} - {{\bar v}_j}|{|^2}}}} \right)^{\frac{1}{{\gamma  - 1}}}}.
\label{eq18}
\end{equation}

Using Eq.~(\ref{eq17}) and Eq.~(\ref{eq18}), we obtain
\begin{equation}
{\left[ {\frac{{{\eta _k}}}{{\lambda \gamma }}} \right]^{\frac{1}{{\gamma  - 1}}}} = {\left( {\sum\limits_j {{{\left[ {\frac{1}{{c_j^\alpha {{({{\widetilde v}_k} - {{\bar v}_j})}^2}}}} \right]}^{\frac{1}{{\gamma  - 1}}}}} } \right)^{ - 1}}.
\label{eq19}
\end{equation}

Substituting Eq.~(\ref{eq19}) in Eq.~(\ref{eq18}), we obtain the updated equation for $R$:

\begin{equation}
\begin{split}
&{r_{kj}} = \frac{{{{(c_j^\alpha ||{{\tilde v}_k} - {{\bar v}_j}|{|^2})}^{\frac{{ - 1}}{{\gamma  - 1}}}}}}{{\sum\nolimits_{{A_t} \ne \emptyset } {{{(c_t^\alpha ||{{\tilde v}_k} - {{\bar v}_t}|{|^2})}^{\frac{{ - 1}}{{\gamma  - 1}}}}} }},\\
&\forall k/{A_k} \subseteq {\Omega _s},{A_k} \ne \emptyset ,\;\;\forall j/{A_j} \subseteq {\Omega _t},{A_j} \ne \emptyset.
\label{eq20}
\end{split}
\end{equation}

\subsubsection{Optimization with respect to cluster centers $V$}
Taking the partial derivative of $J_{TECM}$ to $V$ and setting it to $0$, we obtain
\begin{equation}
\begin{split}
&\frac{{\partial {J_{TECM}}}}{{\partial {v_l}}} = - 2\sum\limits_{i = 1}^n \sum\limits_{{A_{t,j}} \ne \emptyset } c_j^{\alpha  - 1}m_{ij}^\beta {s_{lj}}({x_i} - {{\overline v }_j})\\& - 2\lambda \sum\limits_{{A_{s,k}} \ne \emptyset } {\sum\limits_{{A_{t,j}} \ne \emptyset } {c_j^{\alpha  - 1}r_{kj}^{{\gamma}}{s_{lj}}({{\widetilde v}_k} - {{\overline v }_j})} }
\\&= - 2\sum\limits_{i = 1}^n \sum\limits_{{A_{t,j}} \ne \emptyset } c_j^{\alpha  - 1}m_{ij}^\beta {s_{lj}}({x_i} - \frac{1}{{{c_j}}}\sum\limits_z {{s_{zj}}{v_z}} )\\& - 2\lambda \sum\limits_{{A_{s,k}} \ne \emptyset } {\sum\limits_{{A_{t,j}} \ne \emptyset } {c_j^{\alpha  - 1}r_{kj}^{{\gamma}}{s_{lj}}({{\widetilde v}_k} - \frac{1}{{{c_j}}}\sum\limits_z {{s_{zj}}{v_z}} )} } = 0.
\end{split}
\label{eq21}
\end{equation}
This is equivalent to the following equation:
\begin{equation}
\begin{split}
&\sum\limits_i {{x_i}} \sum\limits_{{A_{t,j}} \ne \emptyset } {c_j^{\alpha  - 1}m_{ij}^\beta {s_{lj}}}  + \lambda \sum\limits_{{A_{s,k}} \ne \emptyset } {{{\widetilde v}_k}} \sum\limits_{{A_{t,j}} \ne \emptyset } {c_j^{\alpha  - 1}r_{kj}^{{\gamma}}{s_{lj}}}
\\&= \sum\limits_z {{v_z}} \sum\limits_i {\sum\limits_{{A_{t,j}} \ne \emptyset } {c_j^{\alpha  - 2}m_{ij}^\beta {s_{lj}}{s_{zj}}} }  + \lambda \sum\limits_z {{v_z}} \sum\limits_{{A_{s,k}} \ne \emptyset } {\sum\limits_{{A_{t,j}} \ne \emptyset } {c_j^{\alpha  - 2}r_{kj}^{{\gamma}}{s_{lj}}{s_{zj}}} }.
\end{split}
\label{eq22}
\end{equation}
Let ${B_{\rm{1}}} \in {R^{c \times p}}$, ${B_{\rm{2}}} \in {R^{c \times p}}$, ${H_{\rm{1}}} \in {R^{c \times c}}$ and ${H_{\rm{2}}} \in {R^{c \times c}}$ be matrices defined as
\begin{equation}
\begin{split}
{B_{{1_{lq}}}} = \sum\limits_{i = 1}^n {{x_{iq}}} \sum\limits_{{A_{t,j}} \ne \emptyset }&{c_j^{\alpha  - 1}m_{ij}^\beta {s_{lj}}}  = \sum\limits_{i = 1}^n {{x_{iq}}} \sum\limits_{{w_l} \in {A_{t,j}}} {c_j^{\alpha  - 1}m_{ij}^\beta}, \\
&l = 1,...,c,\quad q = 1,...,p,
\end{split}
\label{eq23}
\end{equation}

\begin{equation}
\begin{split}
{B_{{2_{lq}}}} = \sum\limits_{{A_{s,k}} \ne \emptyset } {{{\widetilde v}_{kq}}} \sum\limits_{{A_{t,j}} \ne \emptyset }&{c_j^{\alpha  - 1}r_{kj}^{{\gamma}}{s_{lj}}}  = \sum\limits_{{A_{s,k}} \ne \emptyset } {{{\widetilde v}_{kq}}} \sum\limits_{{w_l} \in {A_{t,j}}} {c_j^{\alpha  - 1}r_{kj}^{{\gamma}}}, \\
&l = 1,...,c,\quad q = 1,...,p,
\end{split}
\label{eq24}
\end{equation}

\begin{equation}
\begin{split}
{H_{{1_{lz}}}} = \sum\limits_i \sum\limits_{{A_{t,j}} \ne \emptyset } c_j^{\alpha  - 2}&m_{ij}^\beta {s_{lj}}{s_{zj}}   = \sum\limits_i {\sum\limits_{{A_{t,j}} \supseteq \{ {w_z},{w_l}\} } {c_j^{\alpha  - 2}m_{ij}^\beta } },\\
&l = 1,...,c,\quad z = 1,...,c,
\end{split}
\label{eq25}
\end{equation}

\begin{equation}
\begin{split}
{H_{{2_{lz}}}} = \sum\limits_{{A_{s,k}} \ne \emptyset } \sum\limits_{{A_{t,j}} \ne \emptyset } c_j^{\alpha  - 2}&r_{kj}^{{\gamma}}{s_{lj}}{s_{zj}}   = \sum\limits_{{A_{s,k}} \ne \emptyset } {\sum\limits_{{A_{t,j}} \supseteq \{ {w_z},{w_l}\} } {c_j^{\alpha  - 2}r_{kj}^{{\gamma}}} }, \\
&l = 1,...,c,\quad z = 1,...,c,
\end{split}
\label{eq26}
\end{equation}
where $c$ denotes the number of clusters in the target domain.
Using the above definitions, $V$ is obtained by solving the following linear equation:
\begin{equation}
\begin{split}
{B_1} + \lambda {B_2} = ({H_1} + \lambda {H_2})V.
\end{split}
\label{eq27}
\end{equation}

\subsection{Summary and analysis} \label{4.4}
With the above optimization procedure, the clustering results of the objects can be  finally obtained by the evidential membership $m_{ij}$, $i = 1,...,n$, $\forall j/{A_j} \subseteq {\Omega _t},\;{A_j} \ne \emptyset$, using Eqs. ~(\ref{eq14}) and (\ref{eq15}). The computed evidential membership $M$ creates a credal partition of the objects. The credal partition provides a general clustering framework, which can be reduced to a traditional hard partition, fuzzy partition \cite{1981pattern,membership2020}, possibilistic partition \cite{gargees2020tlpcm,possibilistic1993}, and rough partition \cite{is2015,Dynamic2017}.
A summary of TECM is presented in Algorithm 1.

\begin{algorithm}[htbp]
	\setstretch{1.35}
	\caption{Transfer learning-based evidential c-means (TECM)}
	\label{alg:Framwork}
	\begin{algorithmic}[1]
		\Require
		objects in target domain \{$x_1$,..., $x_n$\}, barycenters learned from source domain $\widetilde v_k$, $\forall {A_k} \subseteq {\Omega _s}, {A_k} \ne \emptyset$, number of clusters $c$, weighting exponents $ \alpha  \geqslant 0$, $\beta  > 1$, ${\gamma} > 1$, distance to the empty set $\delta  > 0$, termination threshold $\varepsilon$, and balance coefficient of transfer learning $\lambda$.
		\Ensure
		credal partition $M$.
		
		\State initialize cluster centers $V(0)$.
		\State $t \gets 0$; $J_{TECM}(0) \gets +\infty $.
		\Repeat
		\State $t \gets t+1$;
		\State compute credal partition $M(t)$ using Eqs.~(\ref{eq14}) and (\ref{eq15});
		\State compute cluster barycenter association matrix $R(t)$ using Eq.~(\ref{eq20});
		\State compute $B_1$, $B_2$, $H_1$, and $H_2$ using Eqs.~(\ref{eq23})-(\ref{eq26});
		\State compute cluster centers $V(t)$ using Eq.~(\ref{eq27});
		\State compute objective function $J_{TECM}(t)$ using Eq.~(\ref{eq6});
		\Until{($|J_{TECM}(t)-J_{TECM}(t-1)|<\varepsilon$)}.
	\end{algorithmic}
\end{algorithm}

\textbf{Generality analysis}:
The proposed TECM can be reduced to the original ECM when $\lambda = 0$ and to TFCM \cite{deng2015transfer} when all the evidential components are constrained to singletons. The proposed TECM provides a fine framework for integrating the advantages of evidential clustering and transfer learning.

\textbf{Convergence analysis}:
The convergence of TECM can be proved by Zangwill's convergence theorem \cite{Hathaway1987} in a similar manner to \cite{Gan2008}. It should be noted that although only a local optimal solution can be obtained for the TECM nonconvex optimization problem, similar to other FCM-like algorithms, this local optimal solution is sufficiently effective in the majority of cases.

\textbf{Complexity analysis}:
Similar to ECM, the computational complexity of TECM is $O(2^c*n)$.
However, in the majority of cases, the assignment of objects to focal sets with high cardinality is less interpretable. Therefore, in real-world scenarios, complexity can be reduced from $O(2^c*n)$ to $O(c^2*n)$ by constraining the focal sets containing at most two clusters.

\begin{remark}
	A similar study was recently presented in \cite{zhou21}, which developed an evidential clustering algorithm based on transfer learning. Compared to this study, our proposed TECM algorithm has several advantages. First, the knowledge learned from the source data is represented by the cluster centers in \cite{zhou21}, whereas in our proposed algorithm, it is represented by the barycenters, which can provide richer class distribution information of the source data. Furthermore, the clustering algorithm in \cite{zhou21} is only applicable to situations where the source and target domains have the same number of clusters, whereas our proposed TECM is also applicable to situations where the number of clusters is different, which has a wider application scope.
\end{remark}

\section{Experiments} \label{five}
In this section, the TECM proposed in this study is evaluated on both synthetic and real datasets. First, the comparison algorithms and indices for the performance evaluation are introduced in Section \ref{5.1}. In Section \ref{5.2} and Section \ref{5.3}, the TECM clustering performance is evaluated using synthetic datasets and texture image datasets and compared with other related algorithms. Finally, in Section \ref{5.4}, a parameter analysis of the proposed TECM is provided.

\subsection{Experimental setup} \label{5.1}
In addition to ECM, six other related algorithms: LSSMTC \cite{2009Learning}, CombKM \cite{2009Learning}, TSC \cite{spectral2012}, TDEM \cite{gaussian2022}, TFCM \cite{deng2015transfer}, and TLEC \cite{zhou21}, are employed for comparison. Among these, LSSMTC and CombKM are representative multitask clustering algorithms, and TSC, TDEM, TFCM, and TLEC are representative transfer-clustering algorithms. The main ideas of these algorithms are described as follows.

$\bullet$ LSSMTC \cite{2009Learning}: The intuitive idea of LSSMTC is to learn a subspace shared by all clustering tasks such that knowledge can be transferred through the learned subspace. The objective of LSSMTC consists of two parts. The first part involves clustering the data of each task individually; the second part involves learning the shared subspace and clustering the data of all the tasks in the shared subspace simultaneously.

$\bullet$ CombKM \cite{2009Learning}: The intuitive idea of CombKM is to perform the c-means algorithm on a combined dataset created by merging the data of all the tasks together.

$\bullet$ TSC \cite{spectral2012}: TSC assumes that embedded feature information can connect two clustering tasks, which can simultaneously improve the clustering performance in two domains. Therefore, the data manifold information of the clustering tasks and feature manifold information shared between the two clustering tasks are both introduced in TSC.

$\bullet$ TDEM \cite{gaussian2022}: TDEM was proposed to improve the clustering performance of distributed gaussian mixture model  clustering and accelerate clustering convergence, where each node is regarded as both the source domain and target domain, which can learn from each other to complete the clustering task in distributed P2P networks.

$\bullet$ TFCM \cite{deng2015transfer}: TFCM is a representative parameter-based transfer-clustering algorithm that uses the cluster centers obtained from the source domain by FCM to guide clustering in the target domain.

$\bullet$ TLEC \cite{zhou21}: Evidential clustering based on transfer learning (TLEC) is another transfer-clustering version of ECM that uses the cluster centers obtained from the source domain by ECM to guide the clustering in the target domain. However, this algorithm requires the same number of clusters in both the source and target domains.

The parameter settings of the involved algorithms are displayed in Table~\ref{tab1}, which are the recommendations of the corresponding authors of each proposal with generally acceptable performance. For the balance coefficient of transfer learning, $\lambda$, the grid search strategy \cite{Deng2010} was used to determine the optimal value. All experiments were implemented using MATLAB 2016b on a computer with an Intel i9-10900K 3.70 GHZ CPU and 64 GB RAM.

\begin{table}[H] \centering
	\centering
	\caption{Parameter settings for involved algorithms}\label{tab1}
	\begin{tabular}{|c|c|}
	\hline
	\textbf{Algorithm}&\textbf{Parameter Settings}\\
	\hline
	ECM \cite{2008ECM} & $\alpha$ = 1, $\beta$ = 2, $\delta$ = 10 \\
	\hline
	& task number $T$ = 2\\
	LSSMTC \cite{2009Learning}  &${l^*} \in \{2, 4, 8, 16\}$\\
	&${\lambda^*} \in \{0.2, 0.4, 0.6, 0.8\}$\\
	\hline
    CombKM \cite{2009Learning}  & $k$ is equal to the number of clusters\\
	\hline
	TSC \cite{spectral2012}  & $\lambda$  = 3, $step$ = 1\\
	\hline
	TDEM \cite{gaussian2022}  &  $\lambda$ $\in$ \{0.5:0.1:1.5\}, $\eta_{1}$=$\eta_{2}$=$10^{-10}$ \\
	\hline
	&The fuzzifers $m_1$, $m_2$ = 2, $\varepsilon {\rm{ = 1}}{{\rm{0}}^{{\rm{ - 3}}}}$    \\
	TFCM \cite{deng2015transfer}& maximum iteration = 100 \\
	& $\lambda$ $\in$ \{0, 0.1, 0.5, 1,5, 10, 50,  100, 300, 500, 1000\}\\
	\hline
    TLEC \cite{zhou21}  & $\alpha$ = 1, $\beta$ = 2, $\delta$ = 10, $\beta_{1}$ = $\beta_{2}$ = 1 \\
    \hline
	& $\alpha$ = 1, $\beta$ = 2, $\delta$ = 10, $\gamma$ = 2, $\varepsilon {\rm{ = 1}}{{\rm{0}}^{{\rm{ - 3}}}}$ \\
	TECM & maximum iteration = 100 \\
	& $\lambda$  $\in$ \{0, 0.1, 0.5, 1,5, 10, 50, 100, 300, 500, 1000\}\\	
	\hline
	\end{tabular}
\end{table}

The clustering performance was evaluated using three popular validity indices \cite{schutze2008introduction}: accuracy ($ac$), rand index ($RI$), and normalized mutual information ($NMI$). Their definitions are as follows.

(1) $ac$
\[ac = \frac{{{n_{cor}}}}{n},\]
where $n$ is the total number of objects and $n_{cor}$ is the number of objects that are clustered correctly. The greater the value of $ac$, the better the clustering performance.

(2) $RI$
\[RI = \frac{{f_{00} + f_{11}}}{{n(n-1)/2}},\]
where $f_{00}$ and $f_{11}$ are the number of any two objects from the same class belonging to different clusters and the same cluster, respectively.
The greater the value of $RI$, the better the clustering performance.

(3) $NMI$
\[NMI(X;Y) = 2\frac{{I(X;Y)}}{{H(X) + H(Y)}},\]
where $H(\cdot)$ is the information entropy and $I(X;Y)$ is the mutual information, with $X$ and $Y$ being the real label and clustering result, respectively. The greater the value of $NMI$, the better the clustering performance.

\subsection{Synthetic dataset test}\label{5.2}
The TECM clustering performance in cases of insufficient target data and sufficient yet contaminated target data is evaluated in this section. In addition, two ambiguous datasets are used to illustrate the advantages of TECM for detecting ambiguous objects.
\subsubsection{Clustering performance evaluation}\label{5.2.1}
In this section, the TECM clustering performance is evaluated on synthetic datasets in cases where the target data are insufficient or sufficient yet contaminated, although an abundance of source data with similar data distributions are available.

\begin{figure}[H]
	\subfigure[dataset S1-1 ($c_s$=3)]{
		\begin{minipage}[t]{0.5\linewidth}
			\centering
			\includegraphics[width=\textwidth]{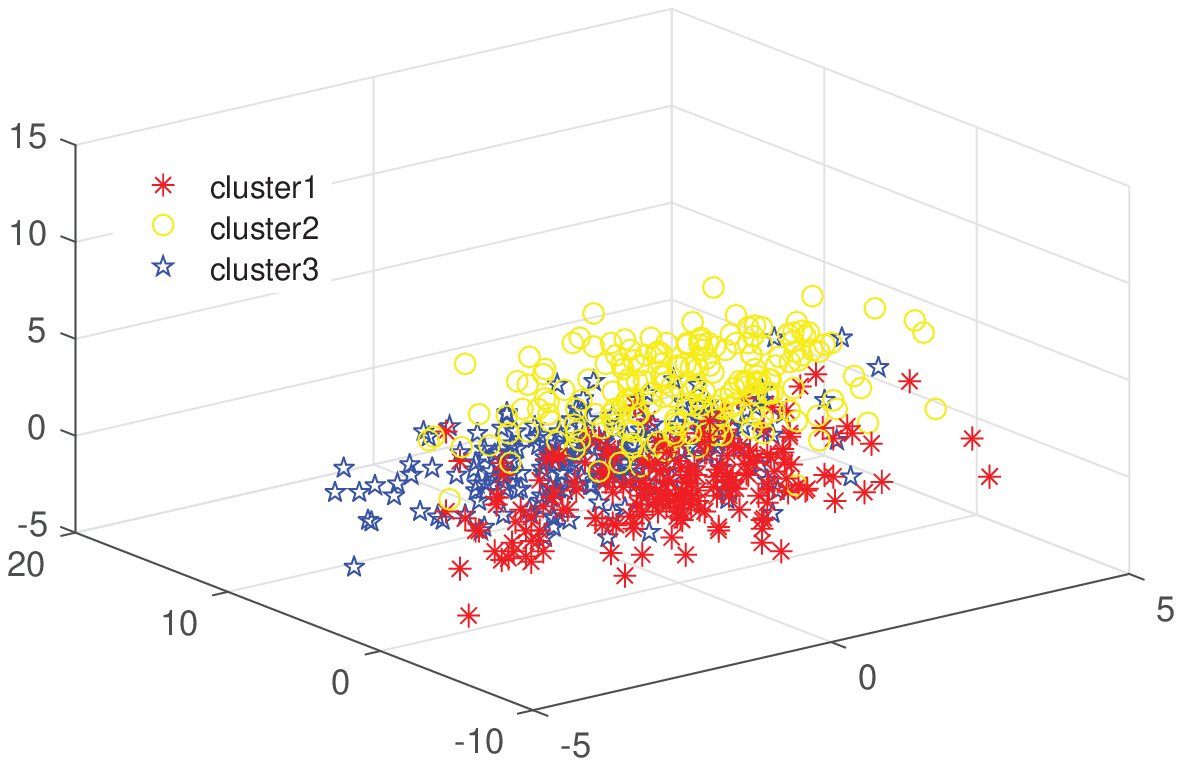}
		\end{minipage}%
	}%
	\subfigure[dataset T1-1 ($c_t$=3)]{
		\begin{minipage}[t]{0.5\linewidth}
			\centering
			\includegraphics[width=\textwidth]{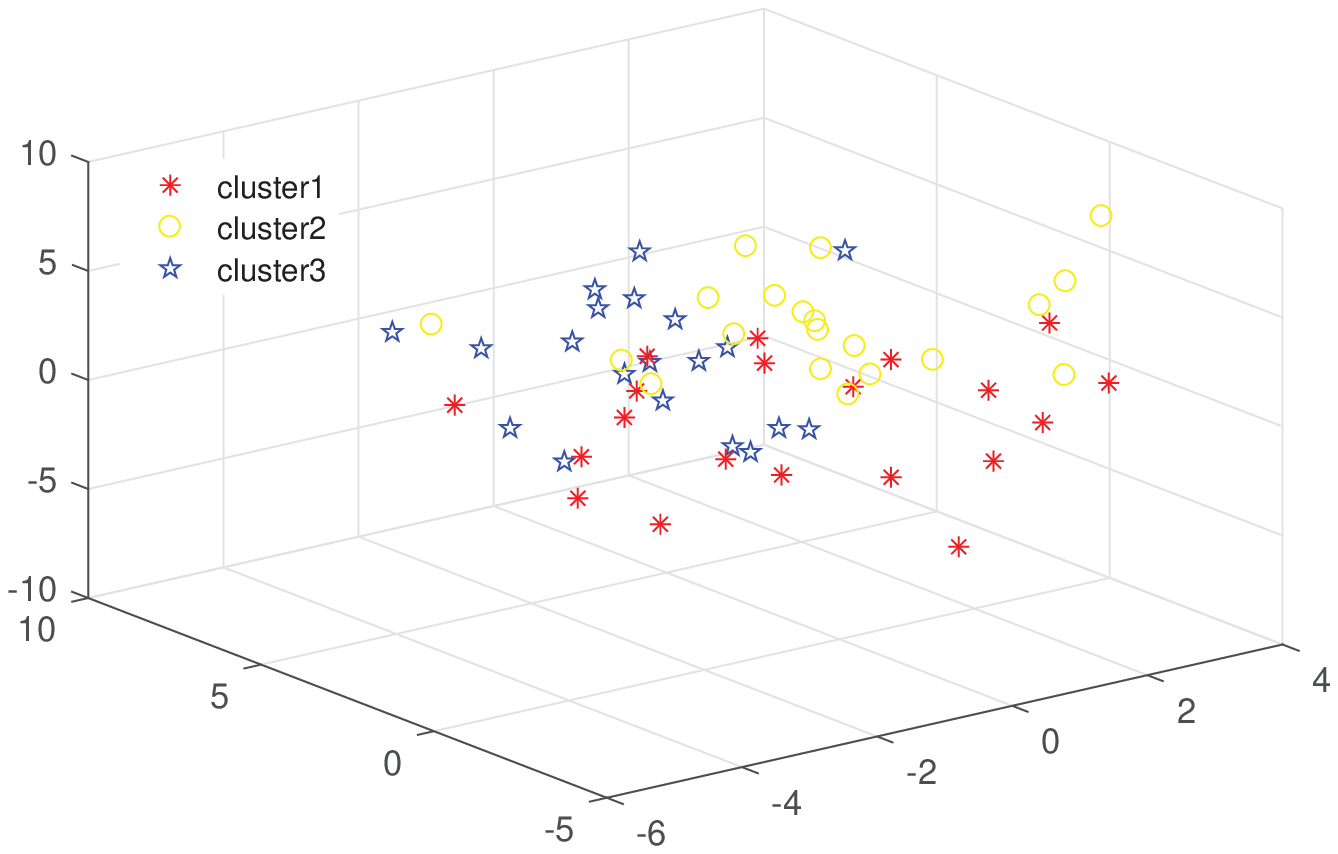}
		\end{minipage}%
	}%
	\\
	\subfigure[dataset S1-2 ($c_s$=4)]{
		\begin{minipage}[t]{0.5\linewidth}
			\centering
			\includegraphics[width=\textwidth]{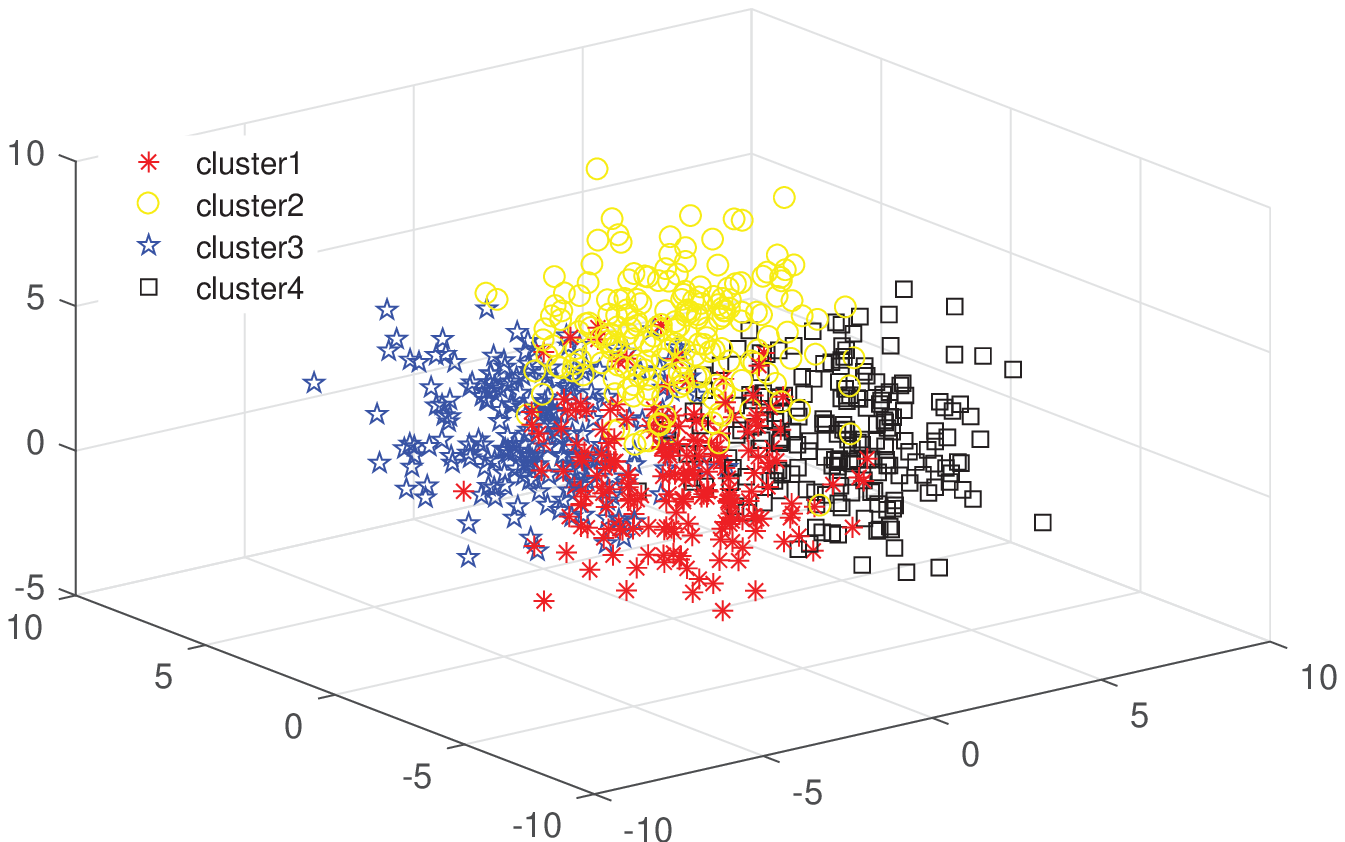}
		\end{minipage}
	}%
	\subfigure[dataset T1-2 ($c_t$=4)]{
		\begin{minipage}[t]{0.5\linewidth}
			\centering
			\includegraphics[width=\textwidth]{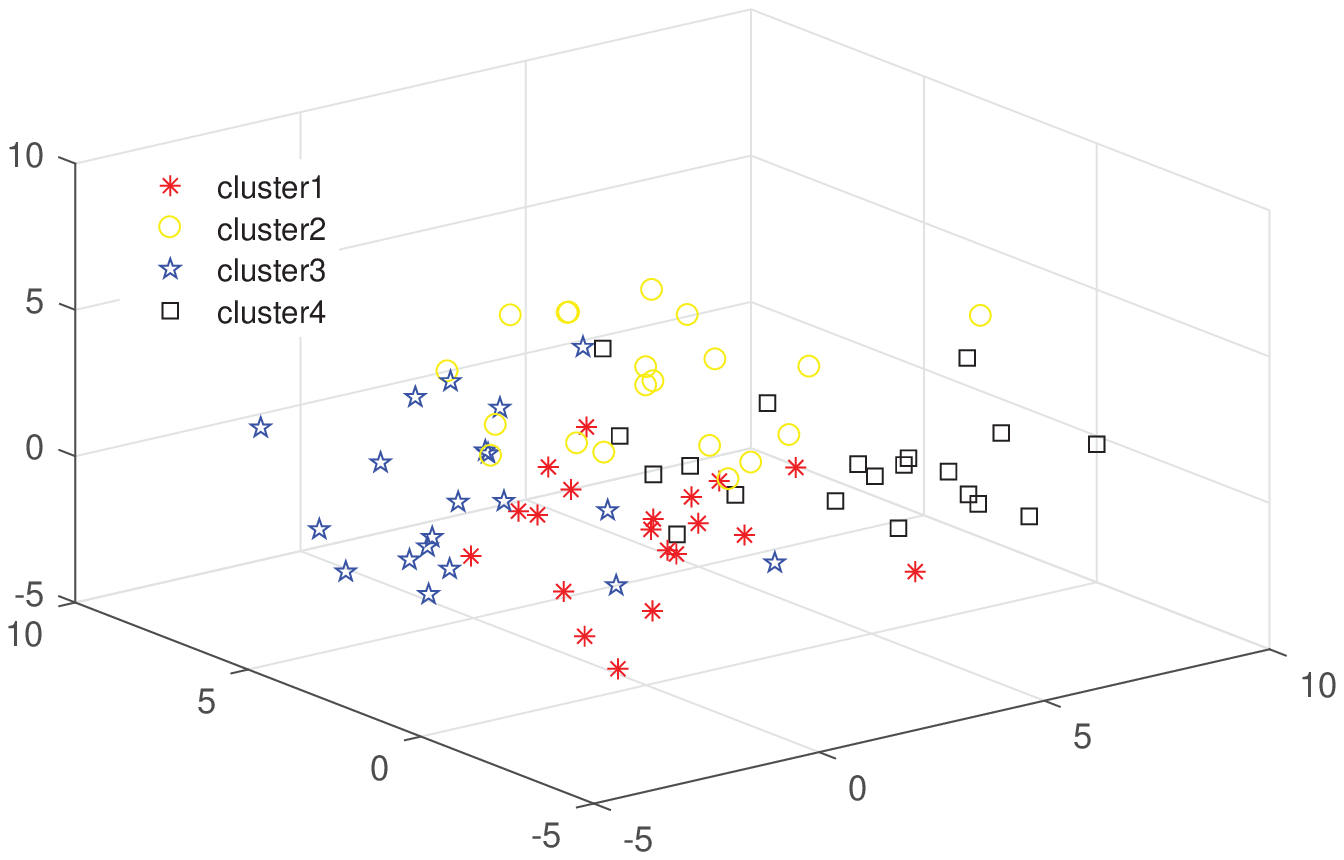}
		\end{minipage}
	}%
	\centering
	\caption{Distributions of synthetic datasets S1-1, T1-1, S1-2, and T1-2.}\label{fig5}
\end{figure}

$\bullet$ \textbf{Insufficient target data}:
Four synthetic datasets, S1-1, T1-1, S1-2, and T1-2, were generated to verify three cases: $c_s=c_t$, $c_s>c_t$, and $c_s<c_t$, where $c_s$ and $c_t$ are the number of clusters in the source and target domains, respectively. The parameters for generating these four synthetic datasets are listed in Table~\ref{tab2}, and the distributions of these datasets are displayed in Fig. ~\ref{fig5}.

\begin{table}[H]
	\centering
	\caption{Parameters for generating synthetic datasets S1-1, T1-1, S1-2, and T1-2}\label{tab2}
	
	\setlength{\tabcolsep}{5mm}
	\begin{tabular}{|c|c|c|c|}
		\hline
		S1-1&mean&covariance&size    \\
		\hline
		Cluster 1&[0 0 0]&[3 0 0; 0 3 0; 0 0 3]&200\\
		Cluster 2&[0 0 5]&[3 0 0; 0 3 0; 0 0 3]&200\\
		Cluster 3&[0 5 0]&[3 0 0; 0 3 0; 0 0 3]&200\\	
	    \hline
		T1-1&mean&covariance&size    \\
		\hline
		Cluster 1&[0 0 0]&[4 0 0; 0 4 0; 0 0 4]&20\\
		Cluster 2&[0 0 5]&[4 0 0; 0 4 0; 0 0 4]&20\\
		Cluster 3&[0 5 0]&[4 0 0; 0 4 0; 0 0 4]&20\\
        \hline
		S1-2&mean&covariance&size\\
		\hline
		Cluster 1&[0 0 0]&[3 0 0; 0 3 0; 0 0 3]&200\\
		Cluster 2&[0 0 5]&[3 0 0; 0 3 0; 0 0 3]&200\\
		Cluster 3&[0 5 0]&[3 0 0; 0 3 0; 0 0 3]&200\\
		Cluster 4&[5 0 0]&[3 0 0; 0 3 0; 0 0 3]&200\\
		\hline
		T1-2&mean&covariance&size\\
		\hline
		Cluster 1&[0 0 0]&[4 0 0; 0 4 0; 0 0 4]&20\\
		Cluster 2&[0 0 5]&[4 0 0; 0 4 0; 0 0 4]&20\\
		Cluster 3&[0 5 0]&[4 0 0; 0 4 0; 0 0 4]&20\\
		Cluster 4&[5 0 0]&[4 0 0; 0 4 0; 0 0 4]&20\\
		\hline
	\end{tabular}
\end{table}

\begin{figure}[H]
	\centering
	\subfigure[dataset T1-3 ($c_t$=3)]{
		\begin{minipage}[t]{0.5\linewidth}
			\centering
			\includegraphics[width=\textwidth]{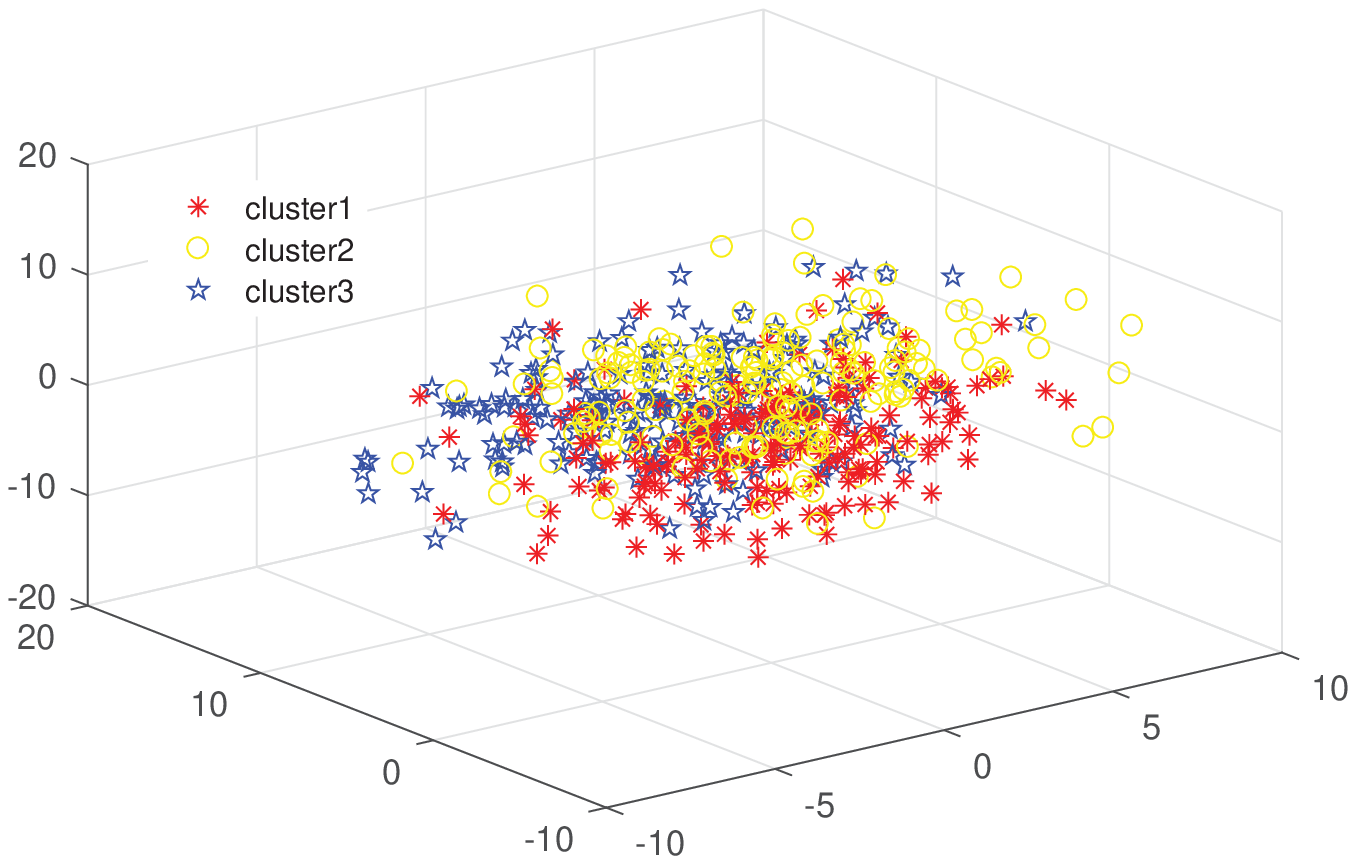}
		\end{minipage}%
	}%
	\subfigure[dataset T1-4 ($c_t$=4)]{
		\begin{minipage}[t]{0.5\linewidth}
			\centering
			\includegraphics[width=\textwidth]{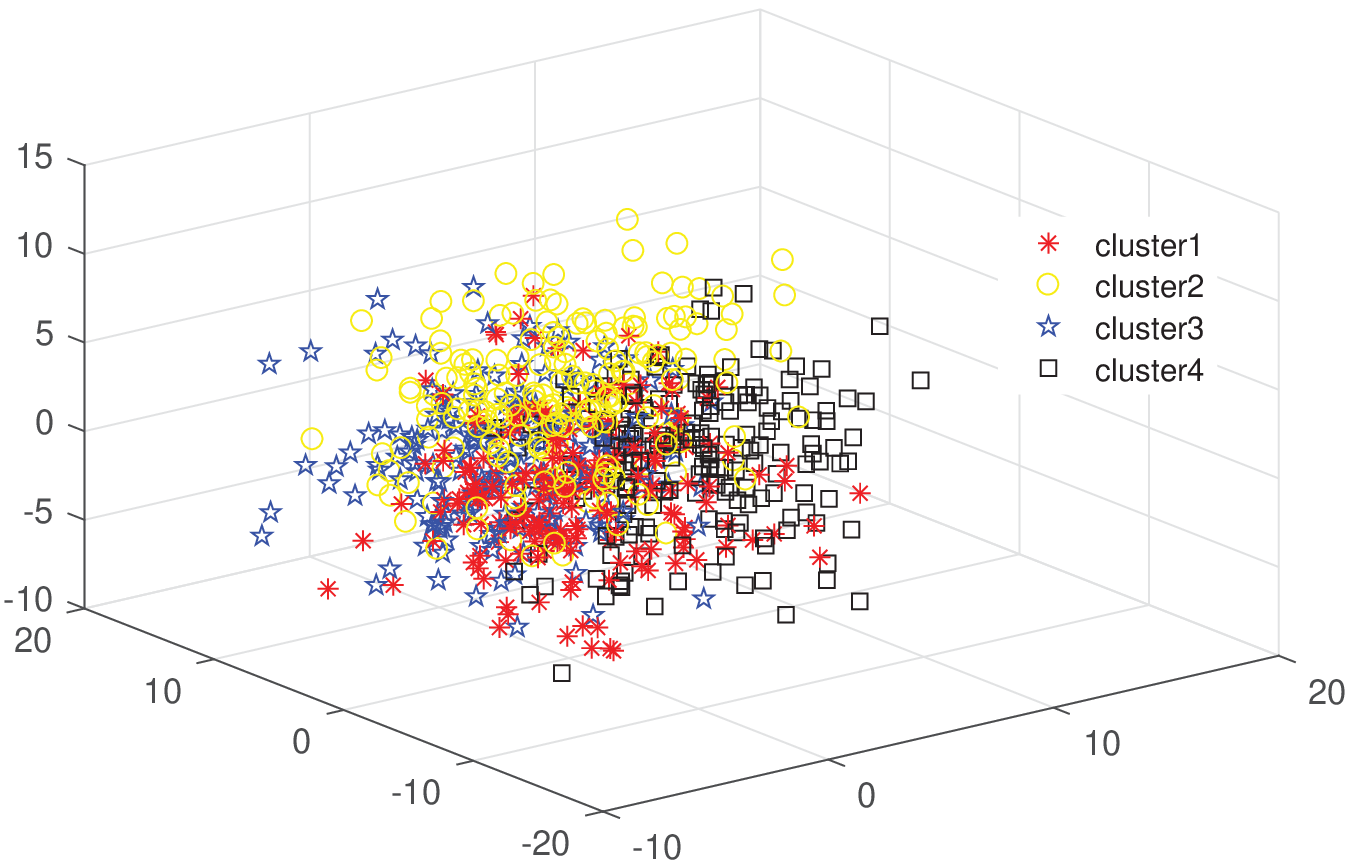}
		\end{minipage}
	}%
	\centering
	\caption{Distributions of synthetic datasets T1-3 and T1-4.}\label{fig6}
\end{figure}

$\bullet$ \textbf{Sufficient yet contaminated target data}:
Two synthetic datasets, T1-3 and T1-4, were generated to evaluate the TECM clustering performance in the case of sufficient yet contaminated target data. Dataset T1-3 was generated by adding zero-mean Gaussian noise with a standard deviation of five to S1-1. Dataset T1-4 was generated by adding zero-mean Gaussian noise with a standard deviation of three to S1-2. The distributions of T1-3 and T1-4 are displayed in Fig. ~\ref{fig6}. Four datasets, S1-1, S1-2, T1-3, and T1-4, were used to verify three cases: $c_s=c_t$, $c_s>c_t$, and $c_s<c_t$.

To minimize the influence of random initialization on the clustering performance evaluation, each algorithm was executed ten times on each target dataset, and then the mean value of each evaluation index was recorded. The clustering performance evaluation results for insufficient target data and sufficient yet contaminated target data are listed in Table~\ref{tab3} and Table~\ref{tab4}, respectively, based on the means of $ac$, $RI$, and $NMI$. In this comparison, ECM used only target data for clustering. LSSMTC and CombKM clustered target data and source data simultaneously, and TSC, TDEM, TFCM, TLEC, and TECM used knowledge learned from the source data to guide the clustering of the target data.
Because LSSMTC, CombKM, TSC, TDEM, and TLEC require that the source and target domains have the same number of clusters, parts of the experimental results are absent.

\begin{table}[H]
	\centering
	\caption{Clustering performance of algorithms for insufficient target data}
	\resizebox{120mm}{28mm}{
		\begin{threeparttable}
			\begin{tabular}{|c|c|c|c|c|c|c|c|c|c|}
				\hline
				\multirow{2}[4]{*}{\textbf{Datasets}} & \multirow{2}[4]{*}{\textbf{Metrics}} & \multicolumn{8}{c|}{\textcolor[rgb]{ .2,  .2,  .2}{\textbf{ Algorithms}}} \bigstrut\\
				\cline{3-10}                 &              & ECM          & LSSMTC       & CombKM       & TSC   &TDEM       & TFCM         & TLEC &TECM \bigstrut\\
				\hline
				\multirow{3}[6]{*}{T1-1 (S1-1$\Rightarrow$T1-1)} & mean\_ac     & 0.850        & 0.670        &0.595         &0.795    &0.767     & \textbf{0.867} & \textbf{0.867} & \textbf{0.867}  \bigstrut\\
				\cline{2-10}                 & mean\_RI     & 0.825        & 0.694        & 0.682        &0.779    &0.763     & \textbf{0.842} & \textbf{0.842} & \textbf{0.842} \bigstrut\\
				\cline{2-10}                 & mean\_NMI    & 0.595        & 0.337        &0.334         &0.471    &0.482     & \textbf{0.626} & \textbf{0.626} & \textbf{0.626}\bigstrut\\
				\hline
				\multirow{3}[6]{*}{T1-1 (S1-2$\Rightarrow$T1-1)} & mean\_ac     & 0.750        & \textbackslash{} & \textbackslash{} & \textbackslash{}    & \textbackslash{}    & 0.850        & \textbackslash{} &\textbf{0.867 } \bigstrut\\
				\cline{2-10}                 & mean\_RI     & 0.757        & \textbackslash{} & \textbackslash{} & \textbackslash{} & \textbackslash{}       & 0.825        & \textbackslash{} &\textbf{0.842 } \bigstrut\\
				\cline{2-10}                 & mean\_NMI    & 0.504        & \textbackslash{} & \textbackslash{} & \textbackslash{}   & \textbackslash{}     & 0.595        & \textbackslash{} &\textbf{0.626 } \bigstrut\\
				\hline
				\multirow{3}[6]{*}{T1-2 (S1-1$\Rightarrow$T1-2)} & mean\_ac     & 0.738        & \textbackslash{} & \textbackslash{} & \textbackslash{} & \textbackslash{} & 0.800        & \textbackslash{} &\textbf{0.813} \bigstrut\\
				\cline{2-10}                 & mean\_RI     & 0.788        & \textbackslash{} & \textbackslash{} & \textbackslash{} & \textbackslash{} & 0.831        & \textbackslash{} &\textbf{0.839} \bigstrut\\
				\cline{2-10}                 & mean\_NMI    & 0.452        & \textbackslash{} & \textbackslash{} & \textbackslash{} & \textbackslash{} & 0.549        & \textbackslash{} &\textbf{0.559} \bigstrut\\
				\hline
			\end{tabular}%
			\begin{tablenotes}
				\footnotesize
				\item[*] For a $\Rightarrow$ b, a and b denote the source and target datasets, respectively.
				
			\end{tablenotes}
		\end{threeparttable}
	}
	\label{tab3}%
\end{table}%

\begin{table}[H]
	\centering
	\caption{Clustering performance of algorithms for sufficient yet contaminated target data}
	\resizebox{120mm}{28mm}{
		\begin{threeparttable}
			\begin{tabular}{|c|c|c|c|c|c|c|c|c|c|}
				\hline
				\multirow{2}[4]{*}{\textbf{Datasets}} & \multirow{2}[4]{*}{\textbf{Metrics}} & \multicolumn{8}{c|}{\textcolor[rgb]{ .2,  .2,  .2}{\textbf{ Algorithms}}} \bigstrut\\
				\cline{3-10}                 &              & ECM          & LSSMTC       & CombKM       & TSC    &TDEM      & TFCM         & TLEC &TECM \bigstrut\\
				\hline
				\multirow{3}[6]{*}{T1-3 (S1-1$\Rightarrow$T1-3)} & mean\_ac     & 0.682        & 0.533        & 0.500        & 0.603   &0.481     & \textbf{0.702} & 0.692 &0.697 \bigstrut\\
				\cline{2-10}                 & mean\_RI     & 0.683        & 0.613        & 0.603        & 0.636    &0.525    & \textbf{0.696} & 0.688 &0.693 \bigstrut\\
				\cline{2-10}                 & mean\_NMI    & 0.253        & 0.140        & 0.123        & 0.168   &0.109     & \textbf{0.274} & 0.258 &0.267  \bigstrut\\
				\hline
				\multirow{3}[6]{*}{T1-3 (S1-2$\Rightarrow$T1-3)} & mean\_ac     & 0.498        & \textbackslash{} & \textbackslash{} & \textbackslash{}   & \textbackslash{}     & 0.522        & \textbackslash{}  &\textbf{0.548} \bigstrut\\
				\cline{2-10}                 & mean\_RI     & 0.599        & \textbackslash{} & \textbackslash{} & \textbackslash{}   & \textbackslash{}    & 0.604        & \textbackslash{}  &\textbf{0.607} \bigstrut\\
				\cline{2-10}                 & mean\_NMI    & 0.097        & \textbackslash{} & \textbackslash{} & \textbackslash{}    & \textbackslash{}    & 0.102        & \textbackslash{}  &\textbf{0.105} \bigstrut\\
				\hline
				\multirow{3}[6]{*}{T1-4 (S1-1$\Rightarrow$T1-4)} & mean\_ac     & 0.619        & \textbackslash{} & \textbackslash{} & \textbackslash{} & \textbackslash{} & 0.635        & \textbackslash{}  &\textbf{0.641} \bigstrut\\
				\cline{2-10}                 & mean\_RI     & 0.721        & \textbackslash{} & \textbackslash{} & \textbackslash{} & \textbackslash{} & 0.728        & \textbackslash{}  &\textbf{0.732} \bigstrut\\
				\cline{2-10}                 & mean\_NMI    & 0.243        & \textbackslash{} & \textbackslash{} & \textbackslash{} & \textbackslash{} & 0.263        & \textbackslash{}  &\textbf{0.264} \bigstrut\\
				\hline
			\end{tabular}%
			\begin{tablenotes}
				\footnotesize
				\item[*] For a $\Rightarrow$ b, a and b denote the source and target datasets, respectively.
			\end{tablenotes}
		\end{threeparttable}
	}
	\label{tab4}%
\end{table}%

Based on the above six groups of experiments, the following conclusions can be drawn.

(1) For all six groups of experiments, the TECM clustering performance was superior to that of ECM, which indicates that the proposed TECM improved the clustering performance with the assistance of the knowledge learned from the source domain.

(2) In the majority of cases, the TECM clustering performance was superior to that of the other representative multitask or transfer-clustering algorithms, which demonstrates its competitive transfer capability.

(3) Compared with TFCM, the performance improvement of TECM was not evident based on the indices of the hard partition. Nevertheless, the advantage of TECM over TFCM was not only reflected in its hard-partition performance. For certain cases with complex data distributions, TECM with credal partition can provide a deeper insight into the data compared to TFCM. This point will be discussed further in future studies.

\subsubsection{Comparison between TECM and TFCM} \label{5.2.2}
\begin{figure}[H]
	\centering
	\subfigure[dataset S2-1]{
		\begin{minipage}[t]{0.5\linewidth}
			\centering
			\includegraphics[width=\textwidth]{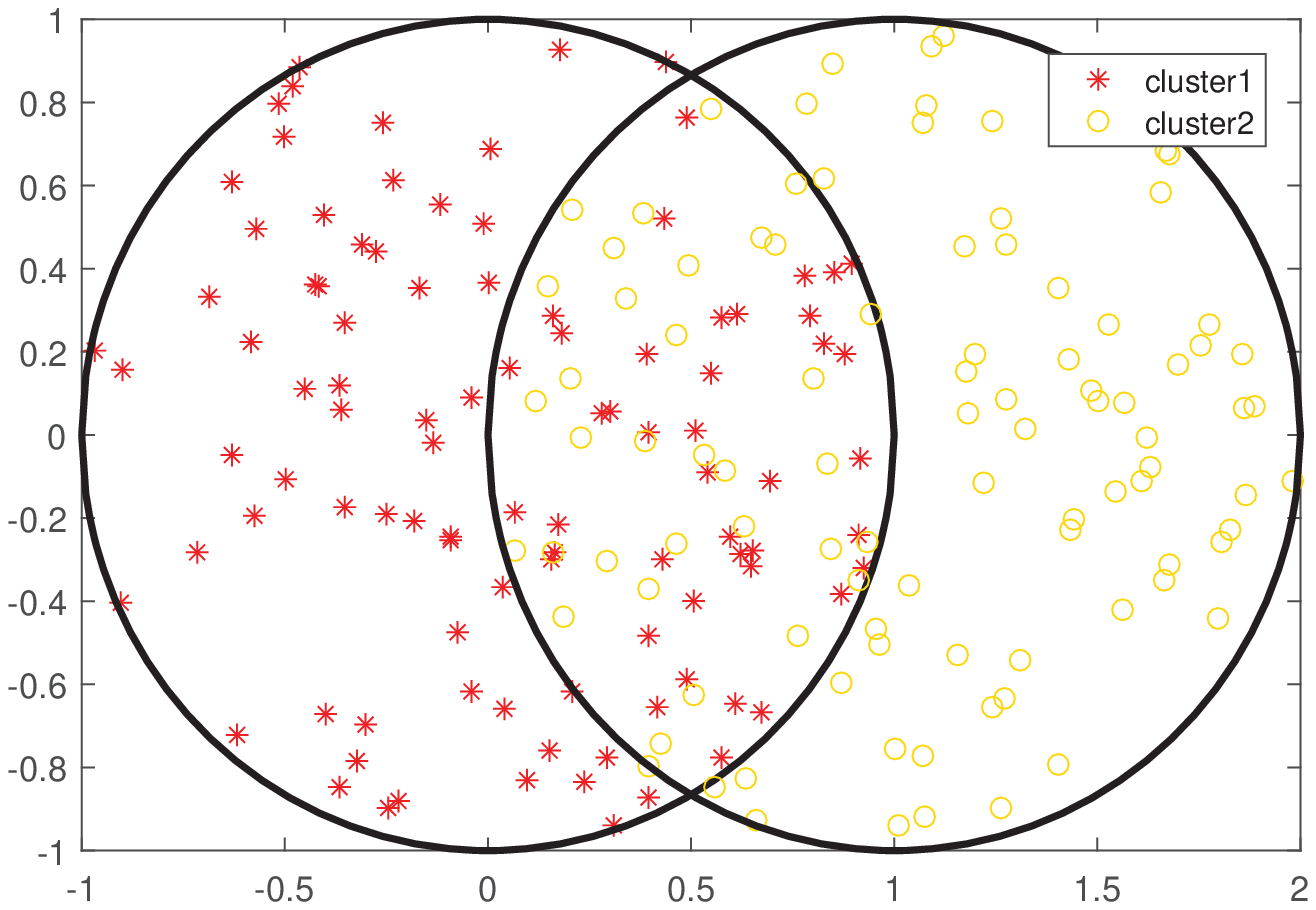}
		\end{minipage}%
	}%
	\subfigure[dataset T2-1]{
		\begin{minipage}[t]{0.5\linewidth}
			\centering
			\includegraphics[width=\textwidth]{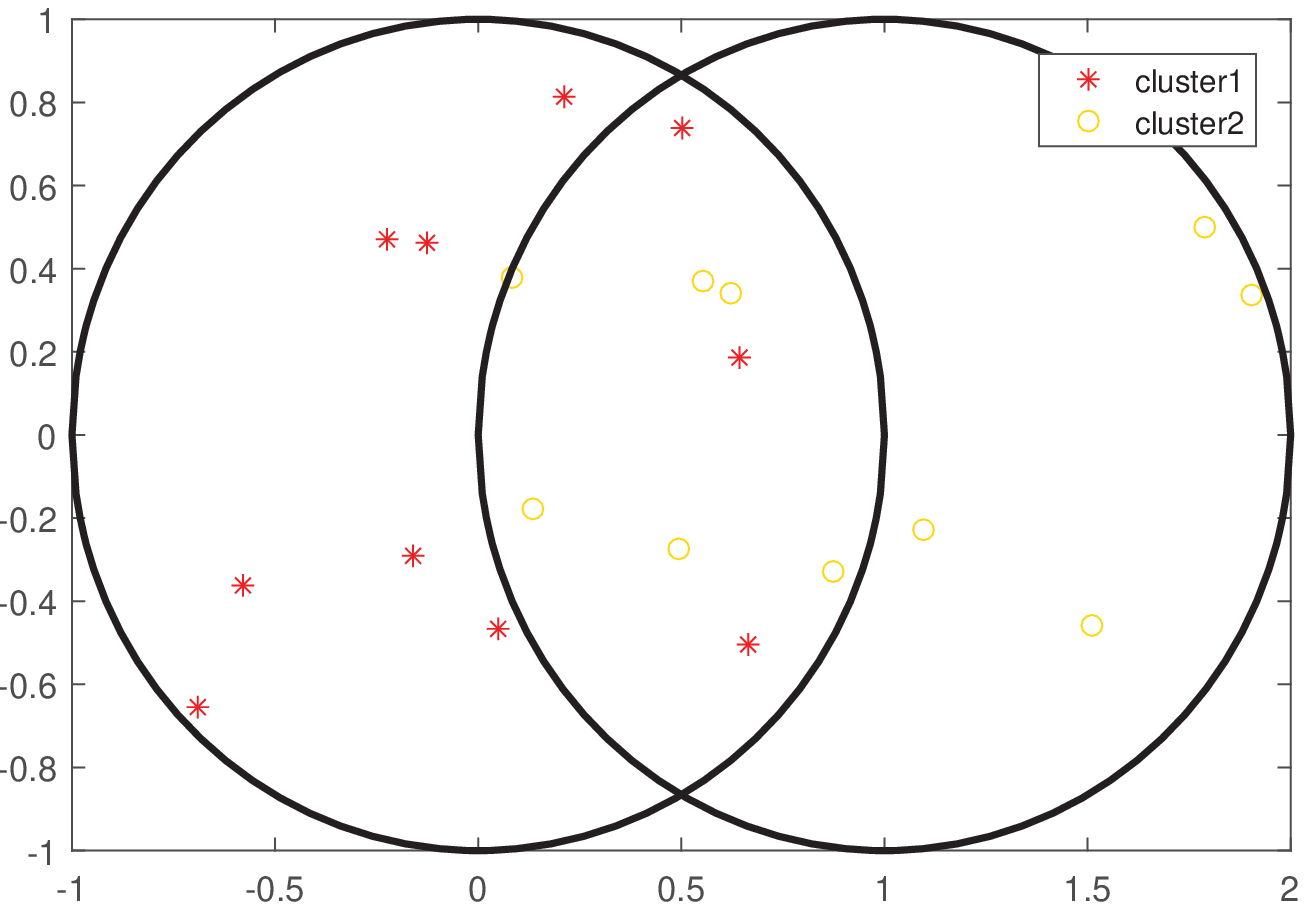}
		\end{minipage}%
	}%
	\\
	\subfigure[dataset S2-2]{
		\begin{minipage}[t]{0.5\linewidth}
			\centering
			\includegraphics[width=\textwidth]{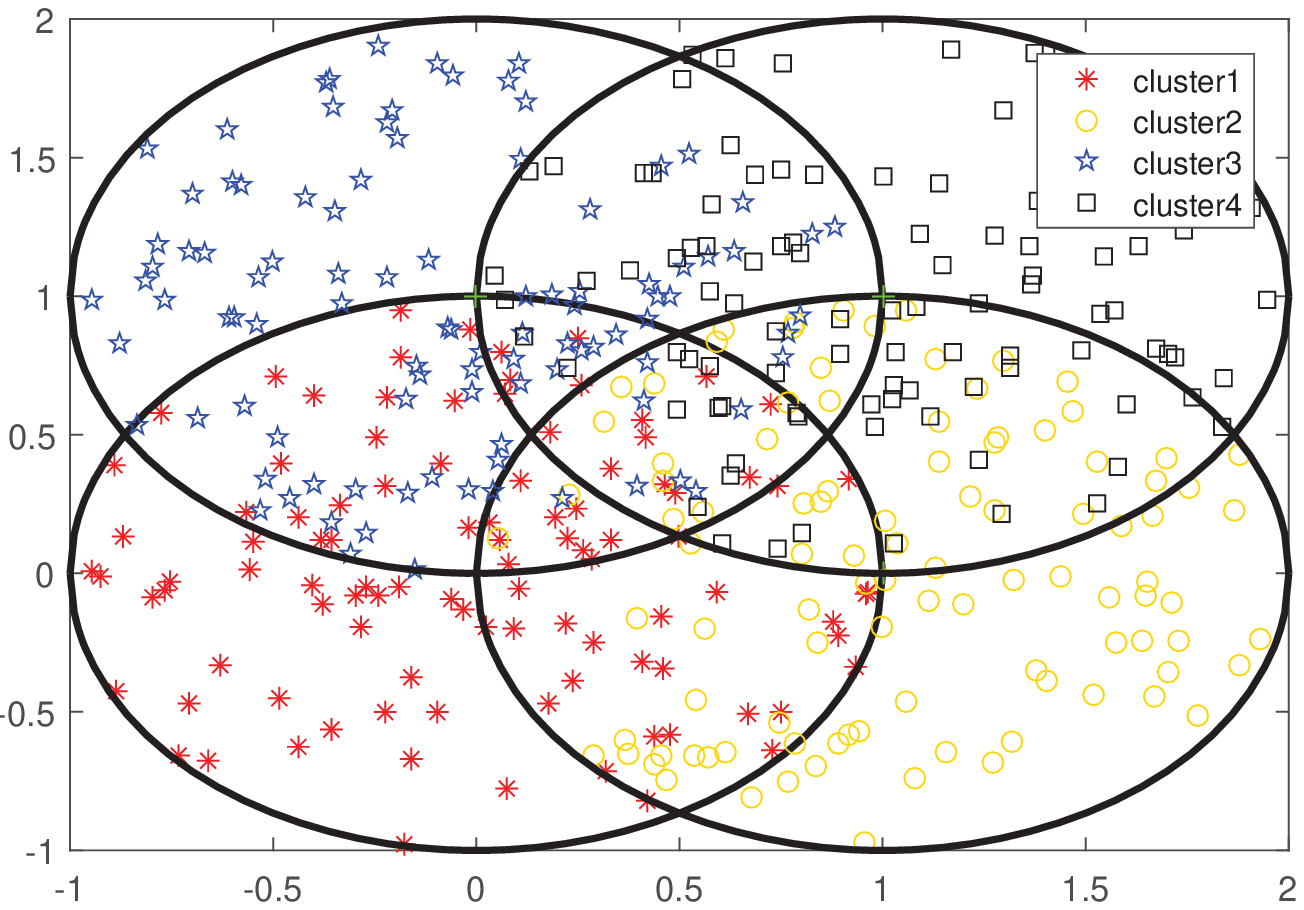}
		\end{minipage}
	}%
	\subfigure[dataset T2-2]{
		\begin{minipage}[t]{0.5\linewidth}
			\centering
			\includegraphics[width=\textwidth]{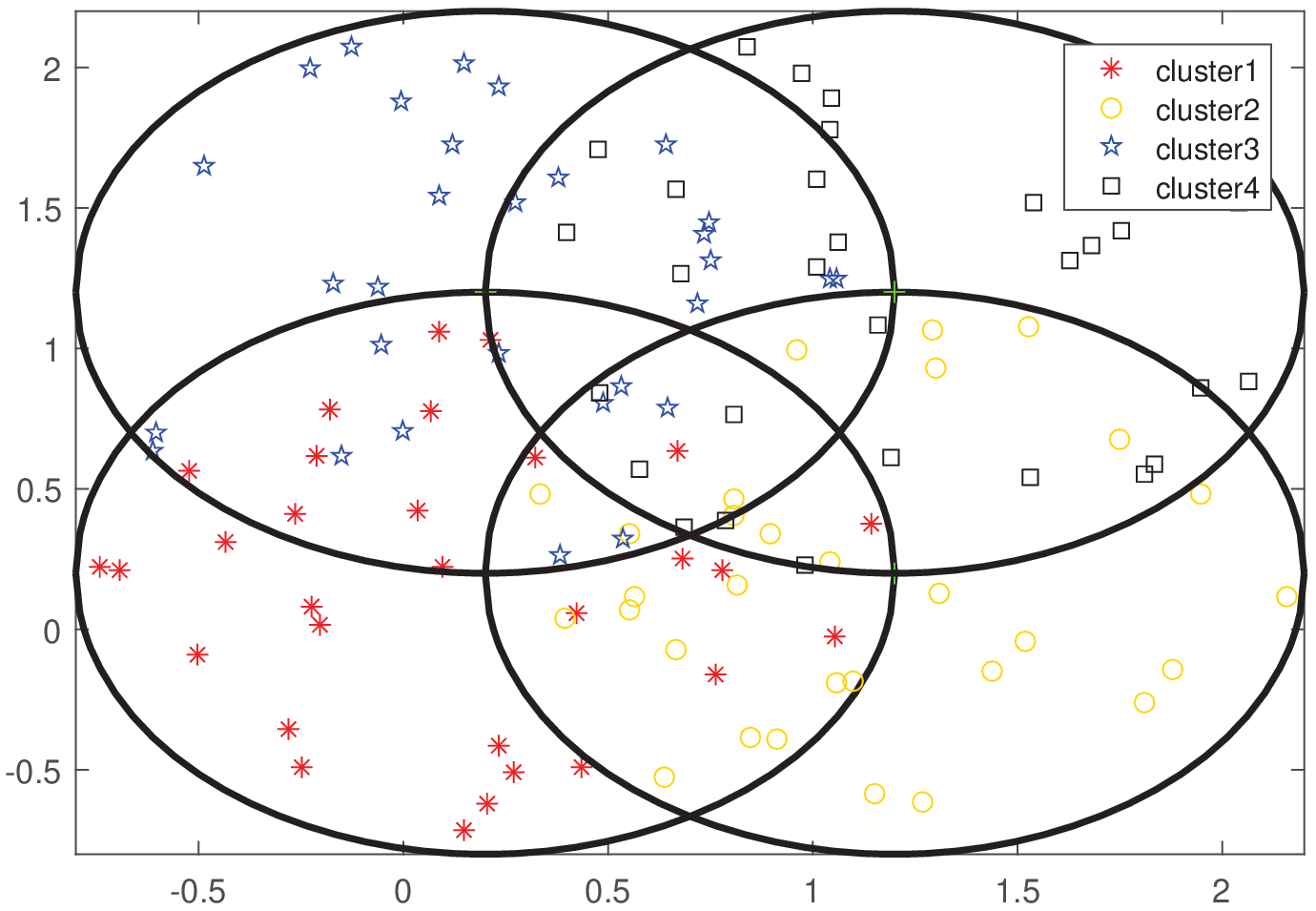}
		\end{minipage}
	}%
	\centering
	\caption{Distributions of synthetic datasets S2-1, T2-1, S2-2, and T2-2.}\label{fig7}
\end{figure}

In this section, two synthetic ambiguous datasets are used to illustrate the superiority of TECM over TFCM in detecting ambiguous objects. The parameters for generating the synthetic datasets T2-1 and T2-2 are listed in Table~\ref{tab5}, and the distributions of the datasets are displayed in Fig. ~\ref{fig7}. The number of clusters in TECM and TFCM were set to two for T2-1 and four for T2-2; TECM had three barycenters for T2-1 ($w_1$, $w_2$ and $w_1 \cup w_2$) and 15 barycenters for T2-2 ($w_1$, $w_2$, $w_1 \cup w_2$,..., $w_1 \cup w_2 \cup w_3 \cup w_4$).
\begin{table}[H]
	\centering
	\caption{Parameters used to generate synthetic datasets S2-1, T2-1, S2-2, and T2-2}\label{tab5}
	\setlength{\tabcolsep}{5mm}
	\resizebox{90mm}{35mm}{
	\begin{tabular}{|c|c|c|c|}
		\hline
		S2-1&mean&covariance&size    \\
		\hline
		Cluster 1&[0 0]&[1 0; 0 1]&100\\
		Cluster 2&[1 0]&[1 0; 0 1]&100\\
		\hline
		T2-1&mean&covariance&size\\
		\hline
        Cluster 1&[0 0.2]&[1 0; 0 1]&10\\
		Cluster 2&[1 0.2]&[1 0; 0 1]&10\\	
		\hline
		S2-2&mean&covariance&size    \\
		\hline
		Cluster 1&[0 0]&[1 0; 0 1]&100\\
		Cluster 2&[1  0]&[1 0; 0 1]&100\\
		Cluster 3&[0  1]&[1 0; 0 1]&100\\
		Cluster 4&[1  1]&[1 0; 0 1]&100\\
		\hline
		T2-2&mean&covariance&size\\
		\hline
        Cluster 1&[0.2 0.2]&[1 0; 0 1]&30\\
		Cluster 2&[1.2 0.2]& [1 0; 0 1]&30\\
		Cluster 3&[0.2 1.2]& [1 0; 0 1]&30\\
		Cluster 4&[1.2 1.2]&[1 0; 0 1]&30\\
		\hline
	\end{tabular}}
\end{table}

\begin{figure}[H]
	\centering
	\subfigure[TFCM for dataset T2-1]{
		\begin{minipage}[t]{0.5\linewidth}
			\centering
			\includegraphics[width=\textwidth]{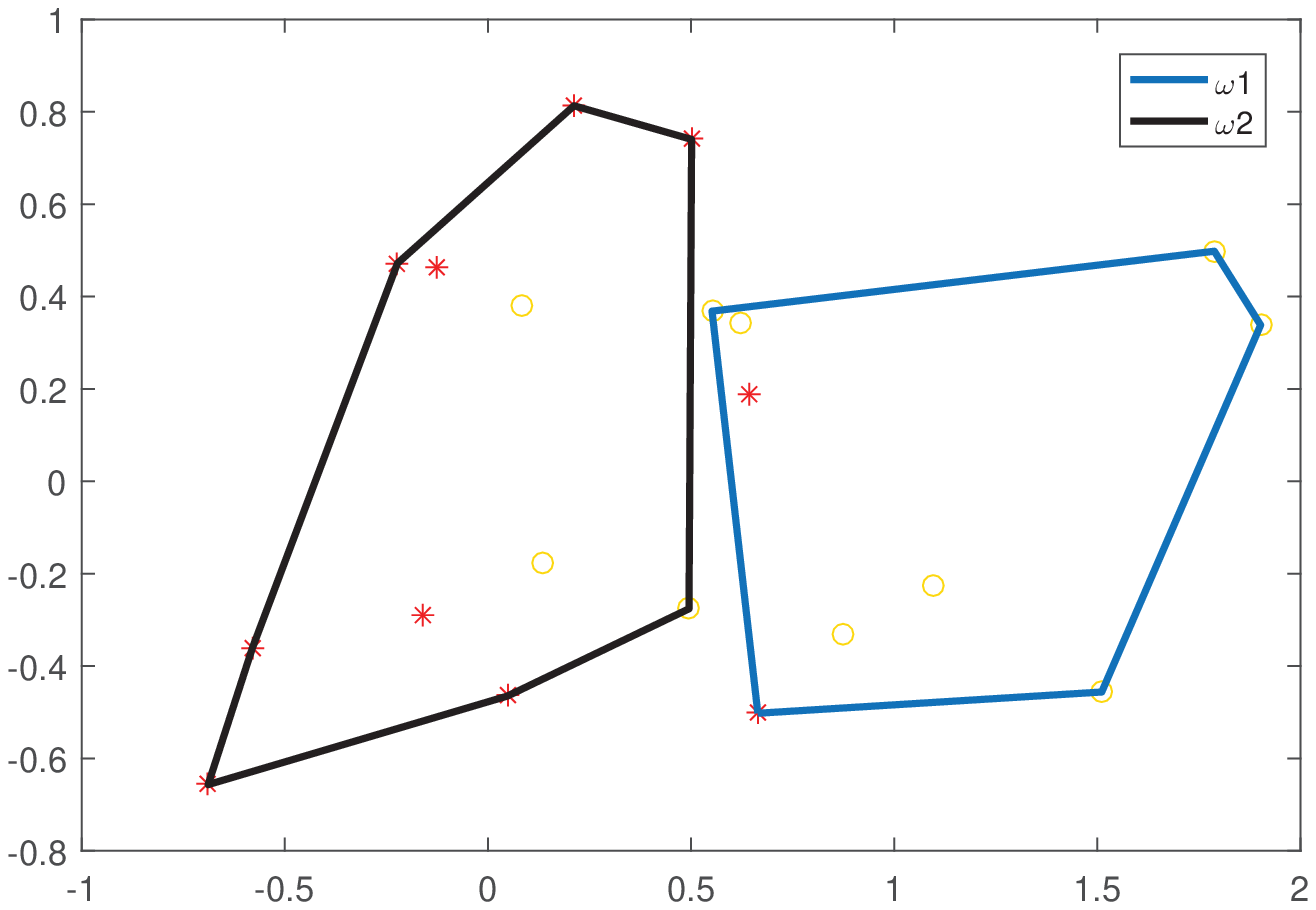}
		\end{minipage}%
	}%
	\subfigure[TECM for dataset T2-1]{
		\begin{minipage}[t]{0.5\linewidth}
			\centering
			\includegraphics[width=\textwidth]{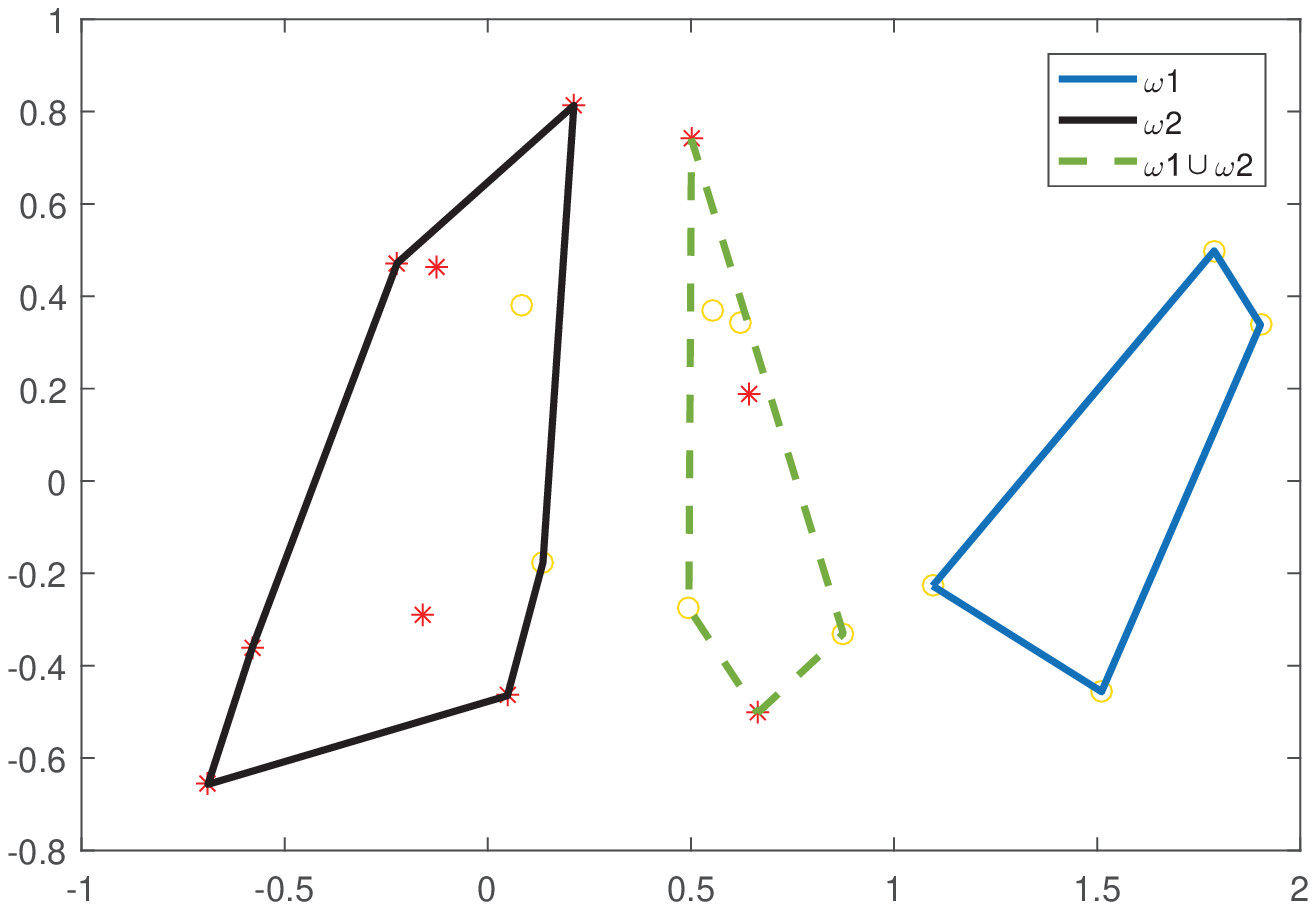}
		\end{minipage}%
	}%
	\\
	\subfigure[TFCM for dataset T2-2]{
		\begin{minipage}[t]{0.5\linewidth}
			\centering
			\includegraphics[width=\textwidth]{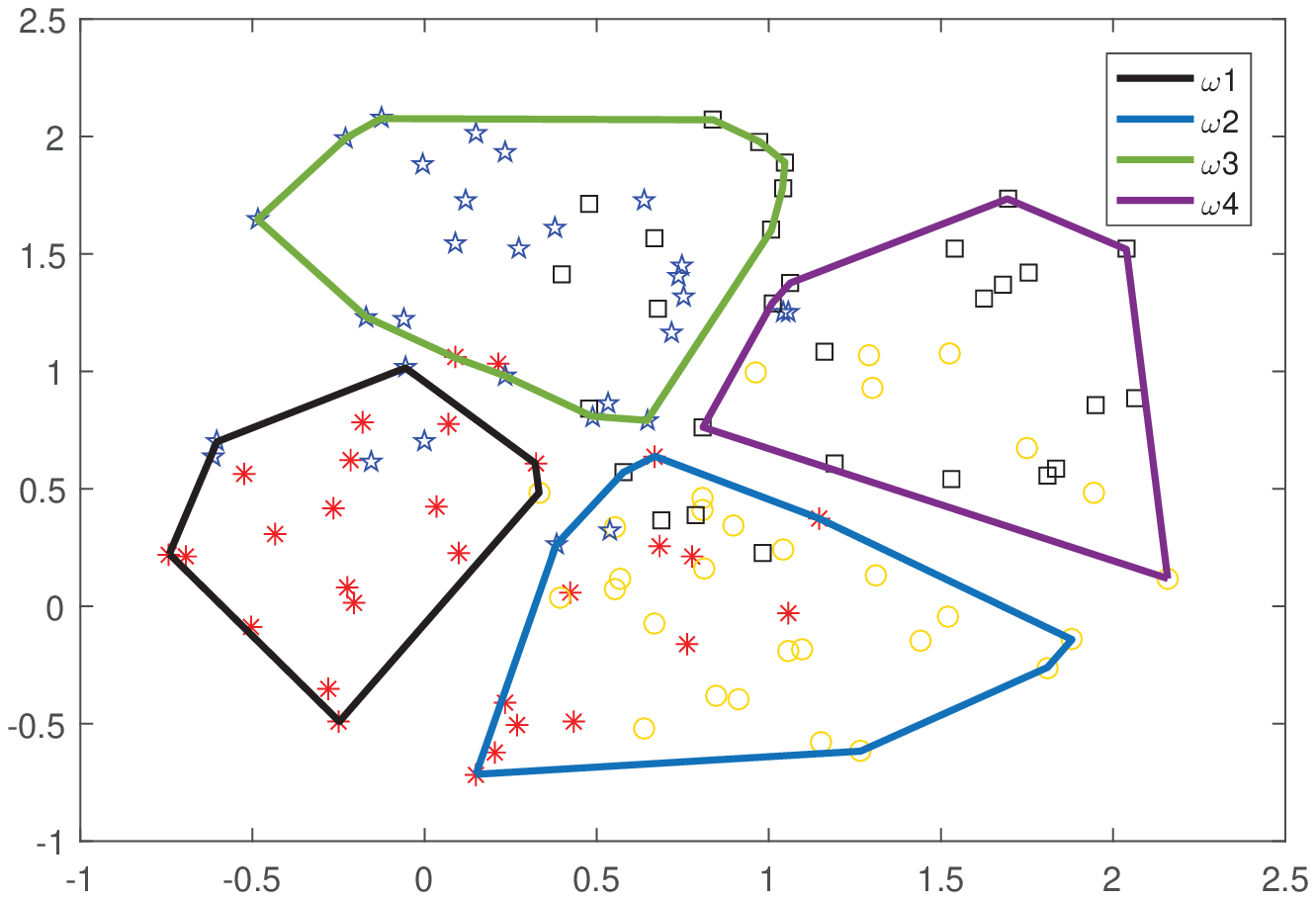}
		\end{minipage}
	}%
	\subfigure[TECM for dataset T2-2]{
		\begin{minipage}[t]{0.53\linewidth}
			\centering
			\includegraphics[width=\textwidth]{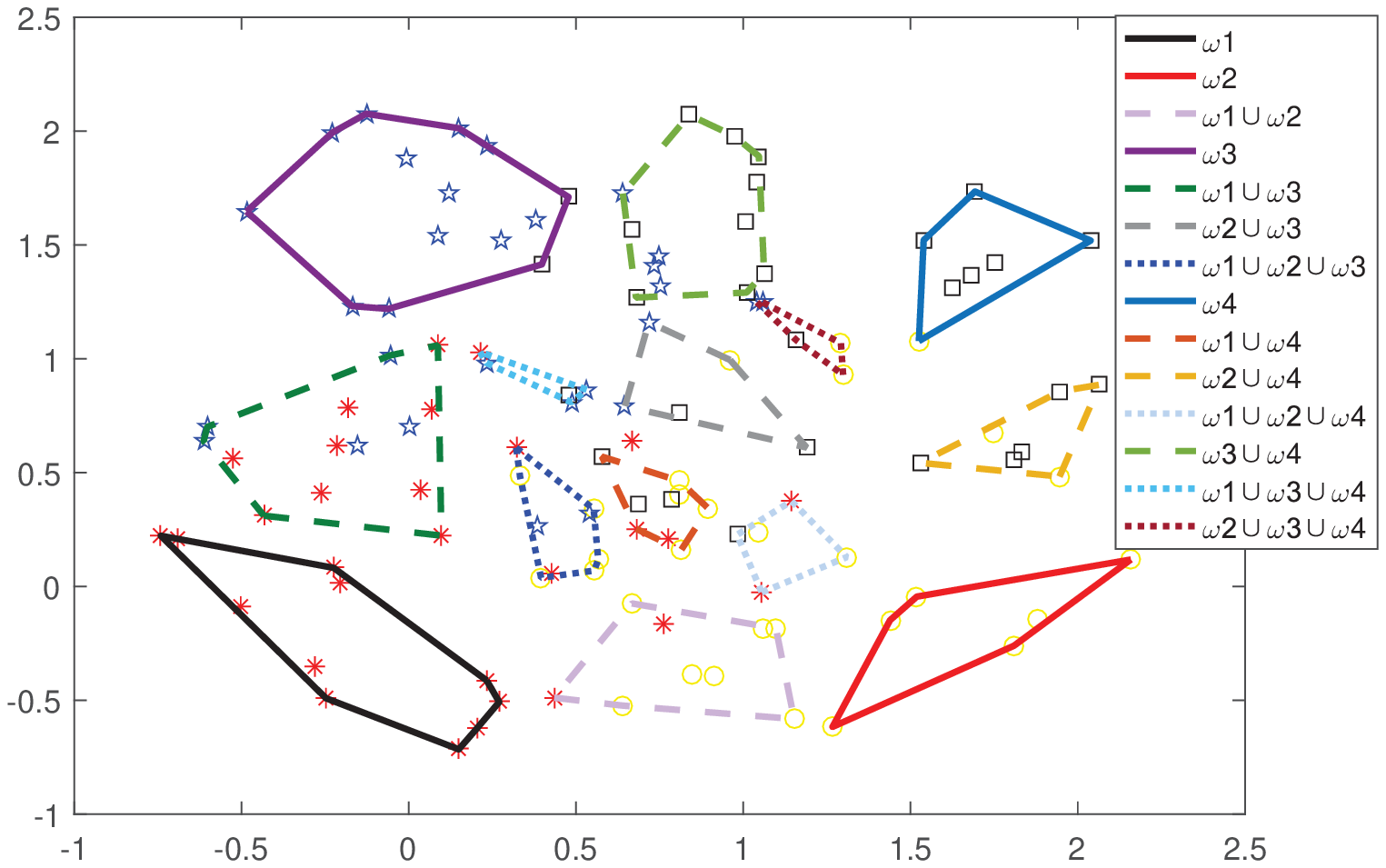}
		\end{minipage}
	}%
	\centering
	\caption{Clustering results of T2-1 and T2-2 using TFCM and TECM.}
	\label{fig8}
\end{figure}

Fig.~\ref{fig8} displays the clustering results of target data T2-1 and T2-2 obtained by transferring the corresponding source data S2-1 and S2-2, respectively. It can be observed that TFCM cannot cluster ambiguous objects well, whereas the proposed TECM can discover those objects located at the boundaries of different clusters and assign them to ambiguous sets of clusters. These examples confirm that the TECM proposed in this study is more effective in detecting ambiguous objects, and thus provides a deeper insight into the distribution of the data. In addition, Table~\ref{tab6} indicates that TECM's clustering performance is superior or comparable to that of TFCM in terms of $ac$, $RI$, and $NMI$.

\begin{table}[H]
	\centering
	\caption{Comparison of clustering performance of TECM and TFCM on T2-1 and T2-2.}
	\small
	\setlength{\tabcolsep}{0.4mm}
	\begin{threeparttable}
		\begin{tabular}{|c|c|c|c|c|c|c|}
			\hline
			\multirow{1}[4]{*}{\textbf{Algorithms}} & \multicolumn{3}{c|}{T2-1 (S2-1$\Rightarrow$T2-1)} & \multicolumn{3}{c|}{T2-2 (S2-2$\Rightarrow$T2-2)} \bigstrut\\
			\cline{2-7}          & mean\_ac    & mean\_RI    & mean\_NMI   & mean\_ac    & mean\_RI    &mean\_NMI \bigstrut\\
			\hline
			TFCM  & 0.850  & 0.732 & 0.399 & 0.633 & 0.735 & 0.371 \bigstrut\\
			\hline
			TECM  & 0.850  & 0.732 & 0.399 & \textbf{0.700}   & \textbf{0.766} & \textbf{0.428} \bigstrut\\
			\hline
		\end{tabular}%
		\begin{tablenotes}
			\footnotesize
			\item[*] For a $\Rightarrow$ b, a and b denote the source and target datasets, respectively.
		\end{tablenotes}
	\end{threeparttable}
	\label{tab6}%
\end{table}%

\subsection{Texture image segmentation}\label{5.3}
\begin{figure}[H]
	\centering
	\subfigure[S3-1 (K=4, $\sigma$=0)]{
		\begin{minipage}[t]{0.3\linewidth}
			\centering
			\includegraphics[width=1in]{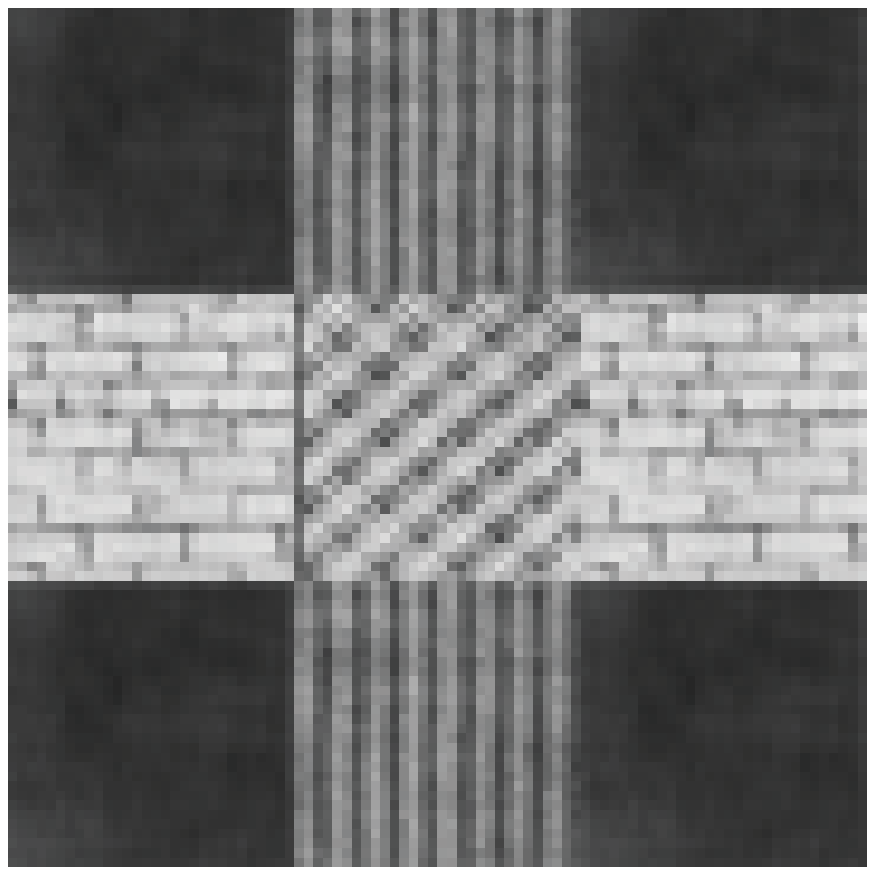}
		\end{minipage}%
	}%
	\quad
	\subfigure[T3-1 (K=4, $\sigma$=0.05)]{
		\begin{minipage}[t]{0.3\linewidth}
			\centering
			\includegraphics[width=1in]{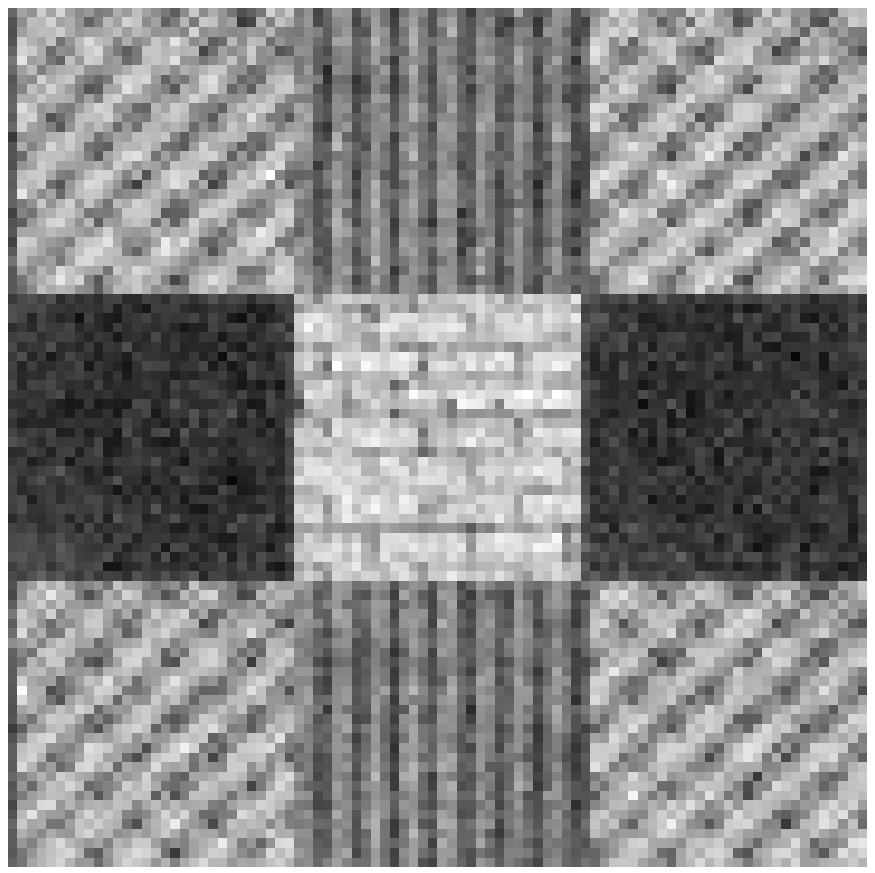}
		\end{minipage}%
	}%
	\subfigure[T3-2 (K=4, $\sigma$=0.1)]{
		\begin{minipage}[t]{0.3\linewidth}
			\centering
			\includegraphics[width=1in]{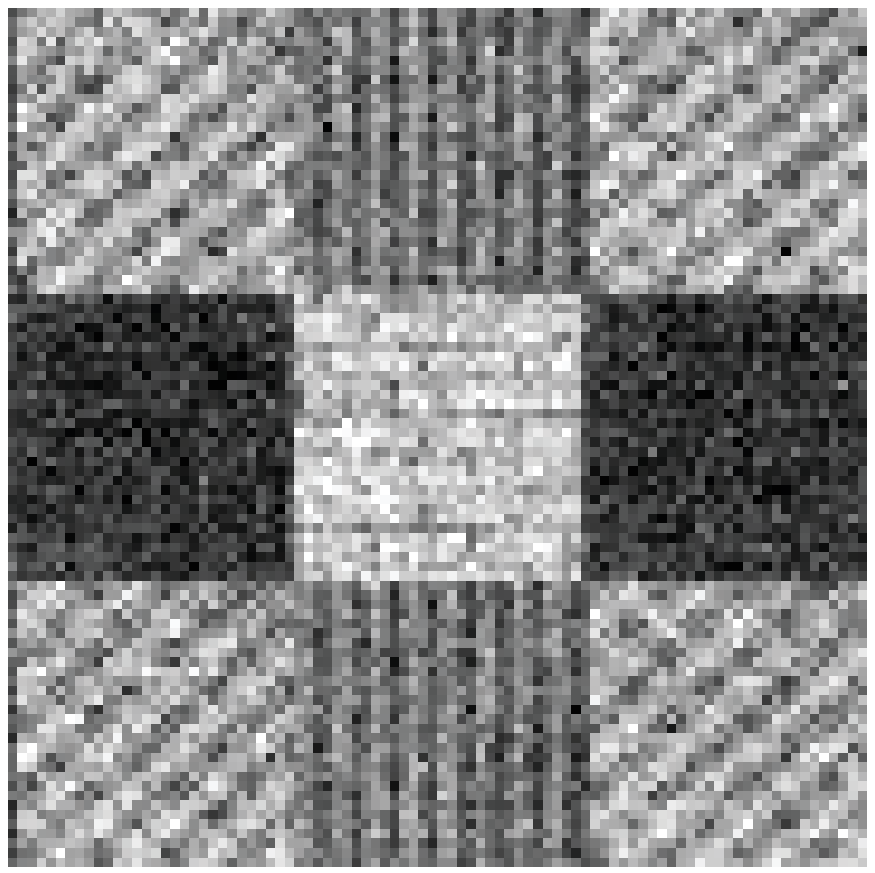}
		\end{minipage}
	}%
	\subfigure[T3-3 (K=3, $\sigma$=0.05)]{
		\begin{minipage}[t]{0.3\linewidth}
			\centering
			\includegraphics[width=1in]{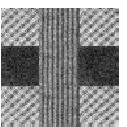}
		\end{minipage}
	}%
	\quad
	\subfigure[T3-4 (K=3, $\sigma$=0.05)]{
		\begin{minipage}[t]{0.3\linewidth}
			\centering
			\includegraphics[width=1in]{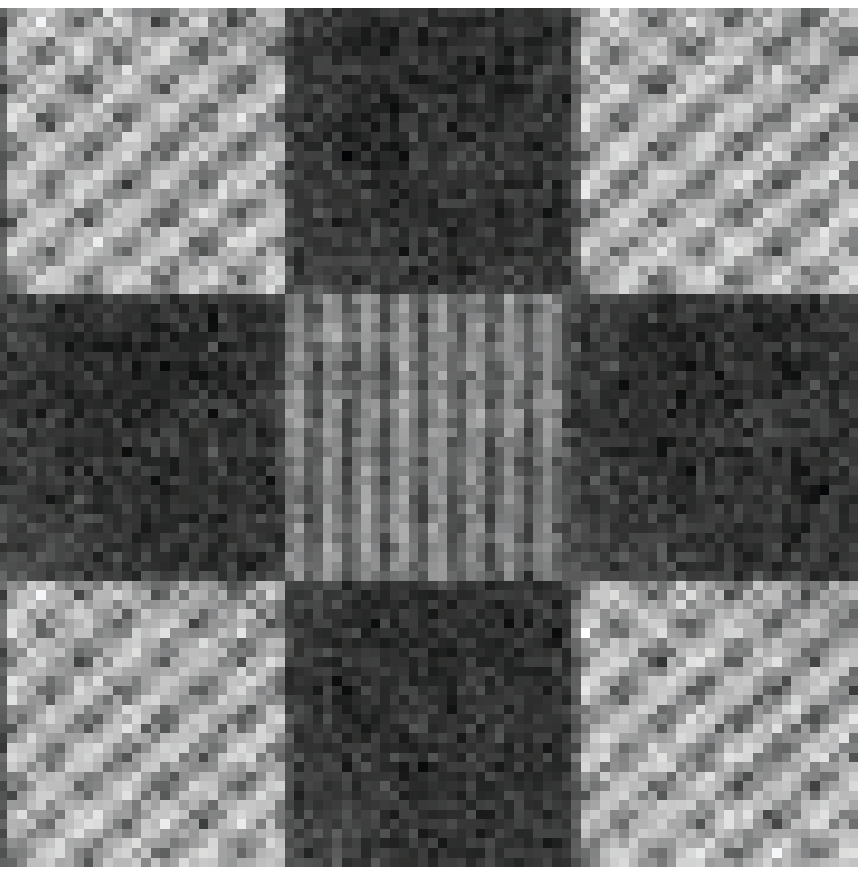}
		\end{minipage}
	}%
	\subfigure[T3-5 (K=5, $\sigma$=0.05)]{
		\begin{minipage}[t]{0.3\linewidth}
			\centering
			\includegraphics[width=1in]{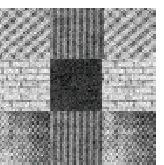}
		\end{minipage}
	}%
	\subfigure[T3-6 (K=5, $\sigma$=0.05)]{
		\begin{minipage}[t]{0.3\linewidth}
			\centering
			\includegraphics[width=1in]{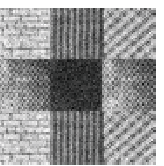}
		\end{minipage}
	}%
	\centering
	\caption{Texture images used in the experiment.}
	\label{fig9}
\end{figure}

\begin{figure}[H]
	\centering
	\subfigure[T3-1]{
		\begin{minipage}[t]{0.3\linewidth}
			\centering
			\includegraphics[width=1in]{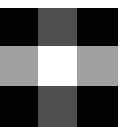}
		\end{minipage}%
	}%
	\subfigure[T3-2]{
		\begin{minipage}[t]{0.3\linewidth}
			\centering
			\includegraphics[width=1in]{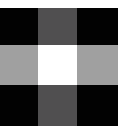}
		\end{minipage}
	}%
	\subfigure[T3-3]{
		\begin{minipage}[t]{0.3\linewidth}
			\centering
			\includegraphics[width=1in]{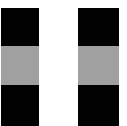}
		\end{minipage}
	}%
	\quad
	\subfigure[T3-4]{
		\begin{minipage}[t]{0.3\linewidth}
			\centering
			\includegraphics[width=1in]{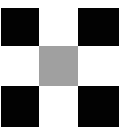}
		\end{minipage}
	}%
	\subfigure[T3-5]{
		\begin{minipage}[t]{0.3\linewidth}
			\centering
			\includegraphics[width=1in]{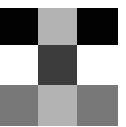}
		\end{minipage}
	}%
	\subfigure[T3-6]{
		\begin{minipage}[t]{0.3\linewidth}
			\centering
			\includegraphics[width=1in]{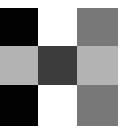}
		\end{minipage}
	}%
	\centering
	\caption{Ideal segmentation of target texture images.}
	\label{fig10}
\end{figure}

In this section, five basic textures D6, D8, D11, D46, and D96 in the Brodatz texture image dataset \cite{Brodatz} are used to create the texture images employed in this experiment; these texture images are then resized to 90 by 90 pixels. In addition, Gaussian noise is added to the target images. Fig.~\ref{fig9} (a) displays the source image and Figs. ~\ref{fig9} (b)--\ref{fig9} (g) display the three types of target images in our experiments. Specifically, Figs.~\ref{fig9} (b) and \ref{fig9} (c) have the same texture number as that in Fig. ~\ref{fig9} (a) but have an inconsistent data distribution. Figs.~\ref{fig9} (d) and \ref{fig9} (e) have smaller texture numbers than those in Fig. ~\ref{fig9} (a)). Figs.~\ref{fig9} (f) and \ref{fig9} (g) have larger texture numbers than those in Fig. ~\ref{fig9} (a)). In addition, the target images had different noise levels. The number of textures and noise levels in each image are given in brackets.
The texture features of the images were obtained using the Gabor filter proposed in \cite{2004Simple}. The ideal segmentation results for the target images are displayed in Fig. ~\ref{fig10}. In each subgraph, small squares of the same color represent the same texture.

\begin{figure}[H]
	\centering
	\subfigure[ECM]{
		\begin{minipage}[t]{0.25\linewidth}
			\centering
			\includegraphics[width=0.98in]{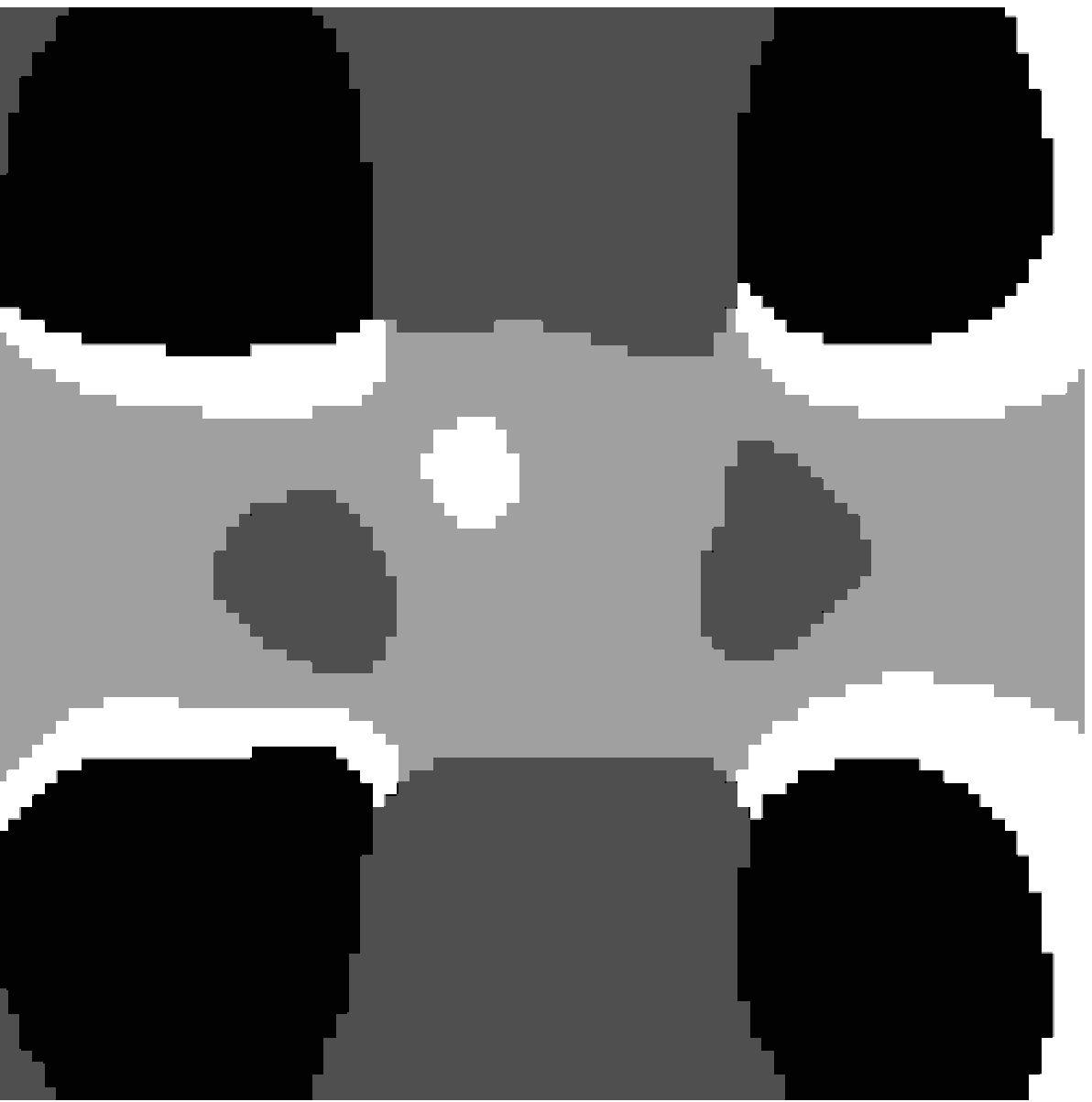}
		\end{minipage}%
	}%
	\subfigure[LSSMTC]{
		\begin{minipage}[t]{0.25\linewidth}
			\centering
			\includegraphics[width=1in]{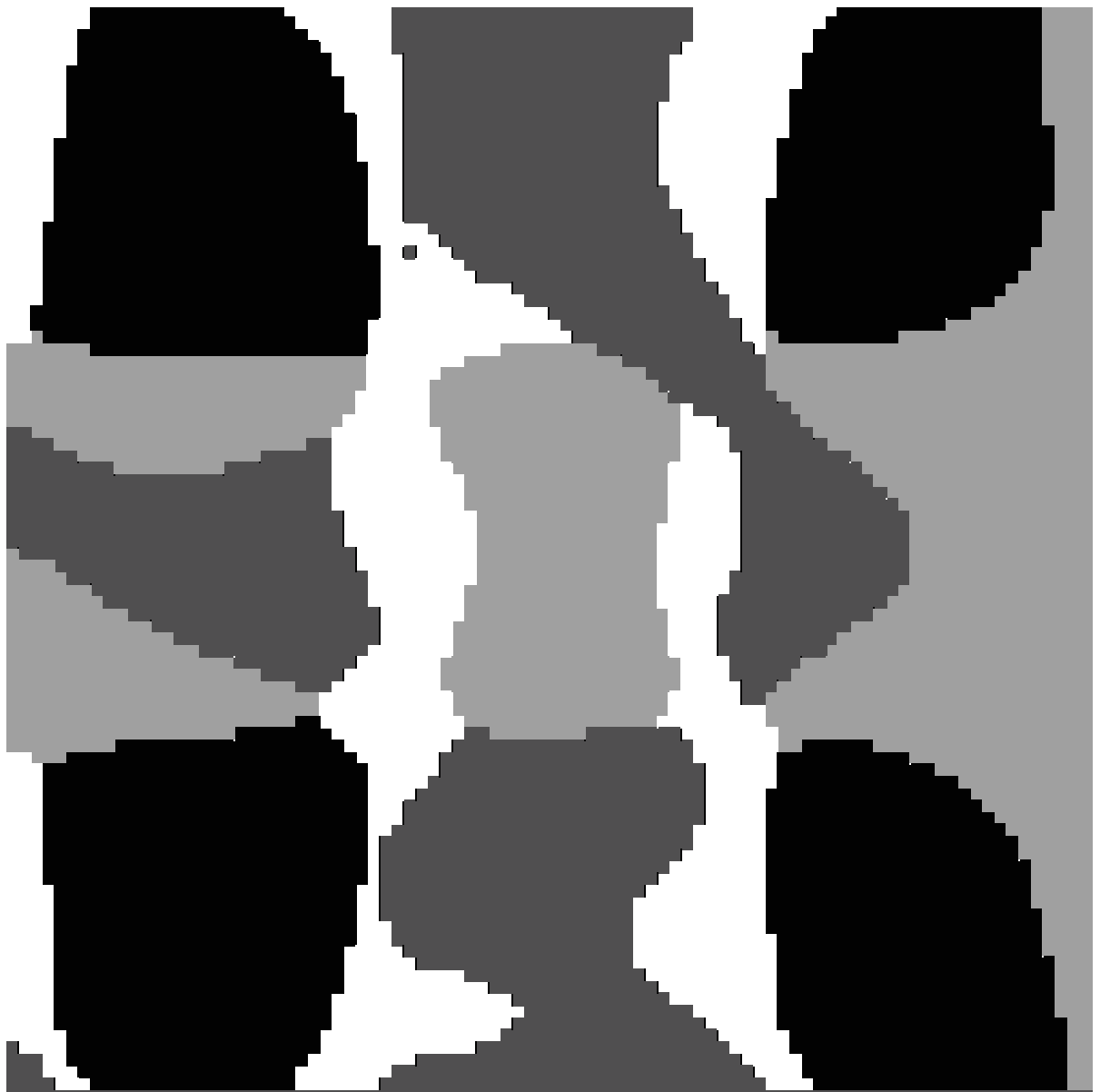}
		\end{minipage}
	}%
	\subfigure[CombKM]{
		\begin{minipage}[t]{0.25\linewidth}
			\centering
			\includegraphics[width=1in]{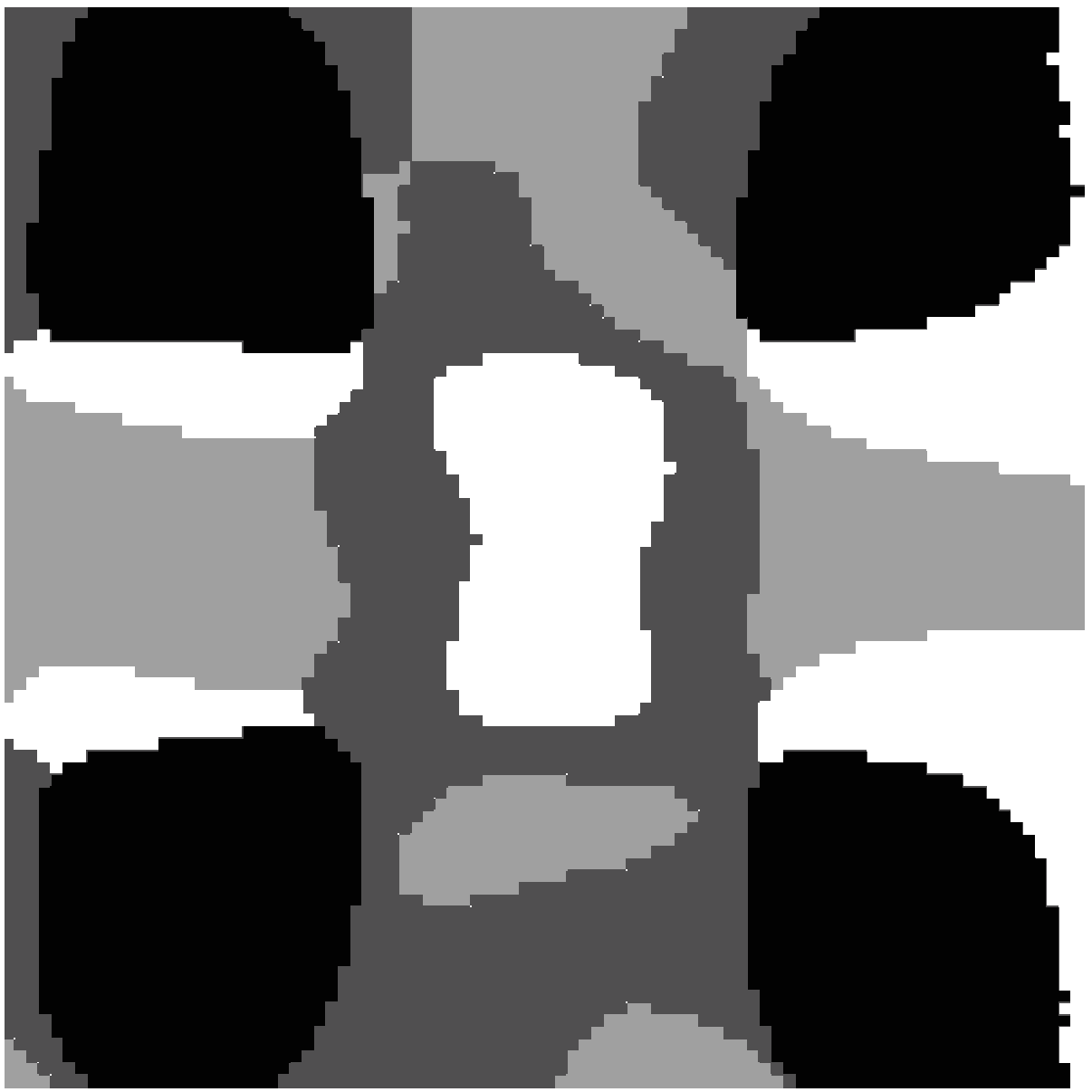}
		\end{minipage}
	}%
	\subfigure[TSC]{
		\begin{minipage}[t]{0.25\linewidth}
			\centering
			\includegraphics[width=1.05in]{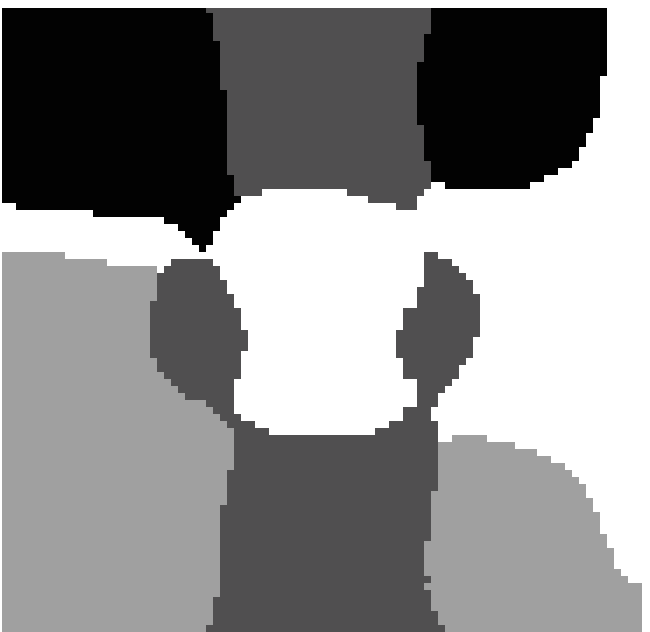}
		\end{minipage}
	}%
 \quad
 \subfigure[TDEM]{
	\begin{minipage}[t]{0.25\linewidth}
		\centering
		\includegraphics[width=1in]{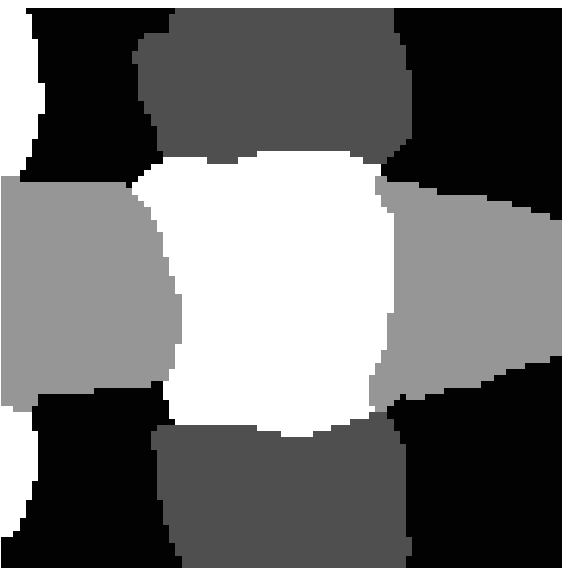}
	\end{minipage}
 }%
	\subfigure[TFCM]{
		\begin{minipage}[t]{0.25\linewidth}
			\centering
			\includegraphics[width=1in]{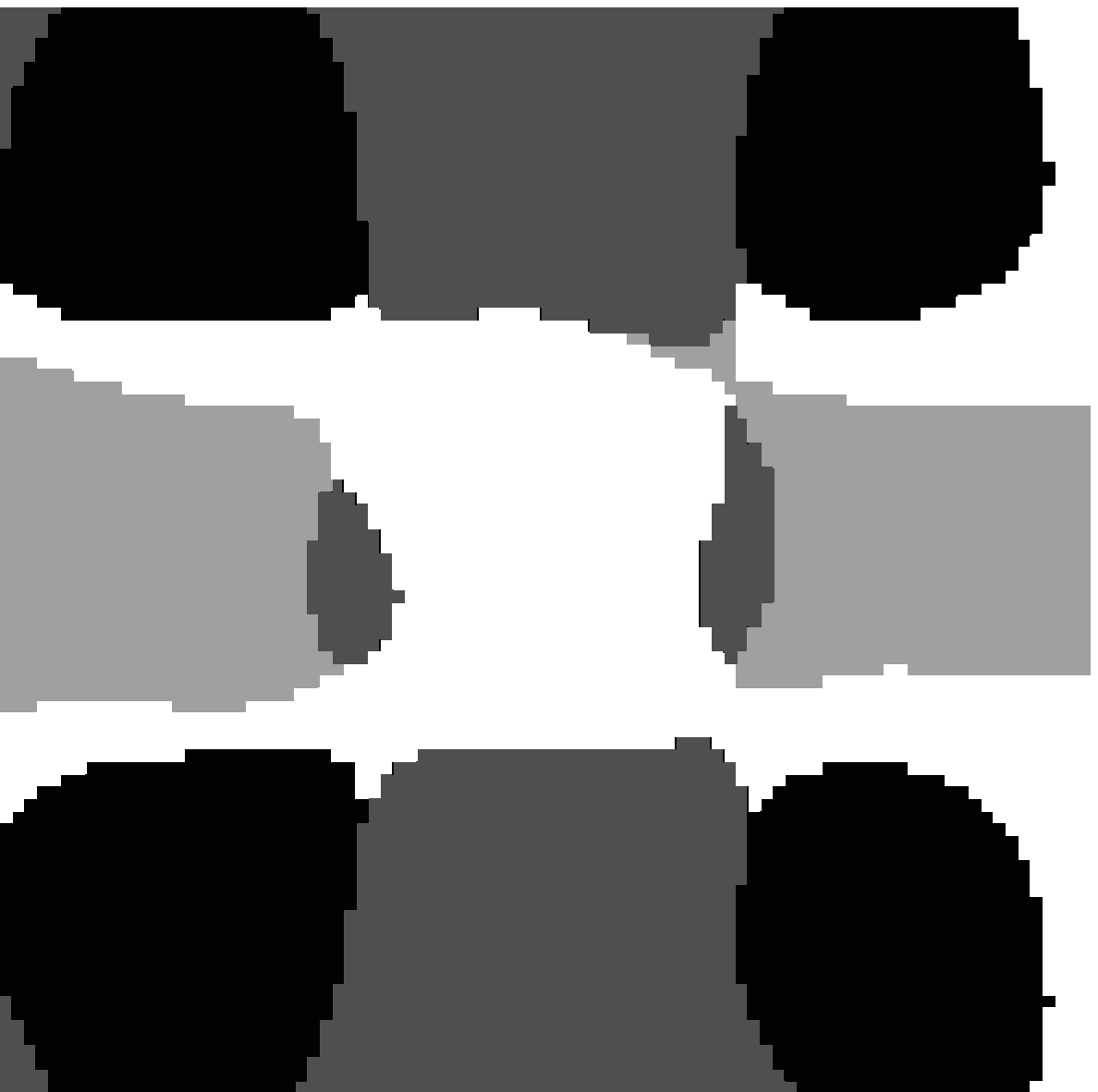}
		\end{minipage}
	}%
\subfigure[TLEC]{
	\begin{minipage}[t]{0.25\linewidth}
		\centering
		\includegraphics[width=1in]{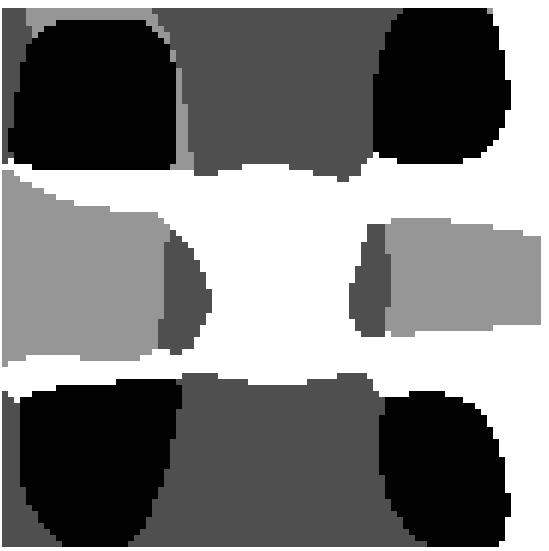}
	\end{minipage}
}%
	\subfigure[TECM]{
		\begin{minipage}[t]{0.25\linewidth}
			\centering
			\includegraphics[width=1in]{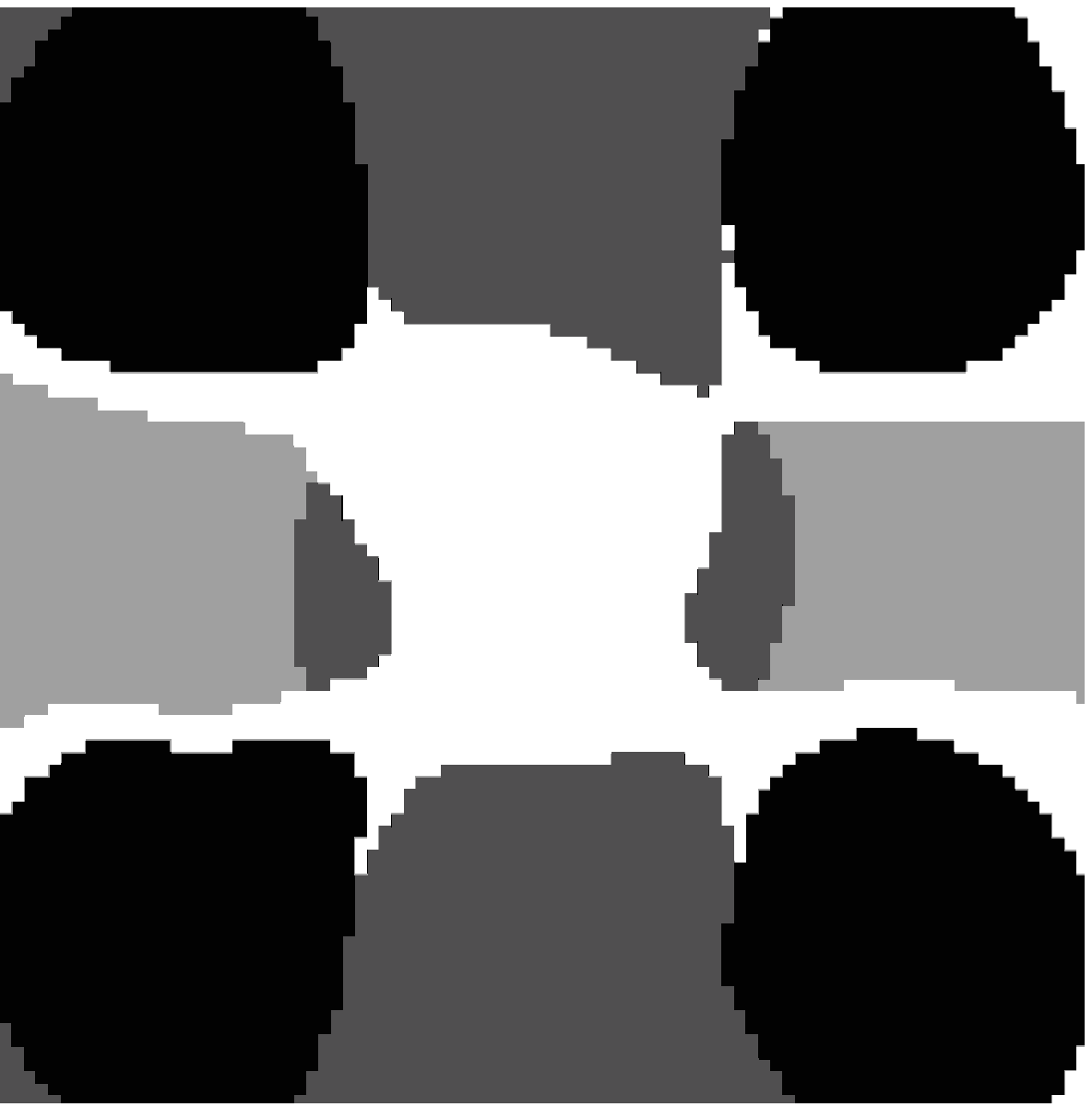}
		\end{minipage}
	}%
	\centering
	\caption{Segmentation results of T3-1 obtained by eight clustering algorithms.}
	\label{fig11}
\end{figure}

\begin{figure}[H]
	\centering
	\subfigure[ECM]{
		\begin{minipage}[t]{0.25\linewidth}
			\centering
			\includegraphics[width=0.98in]{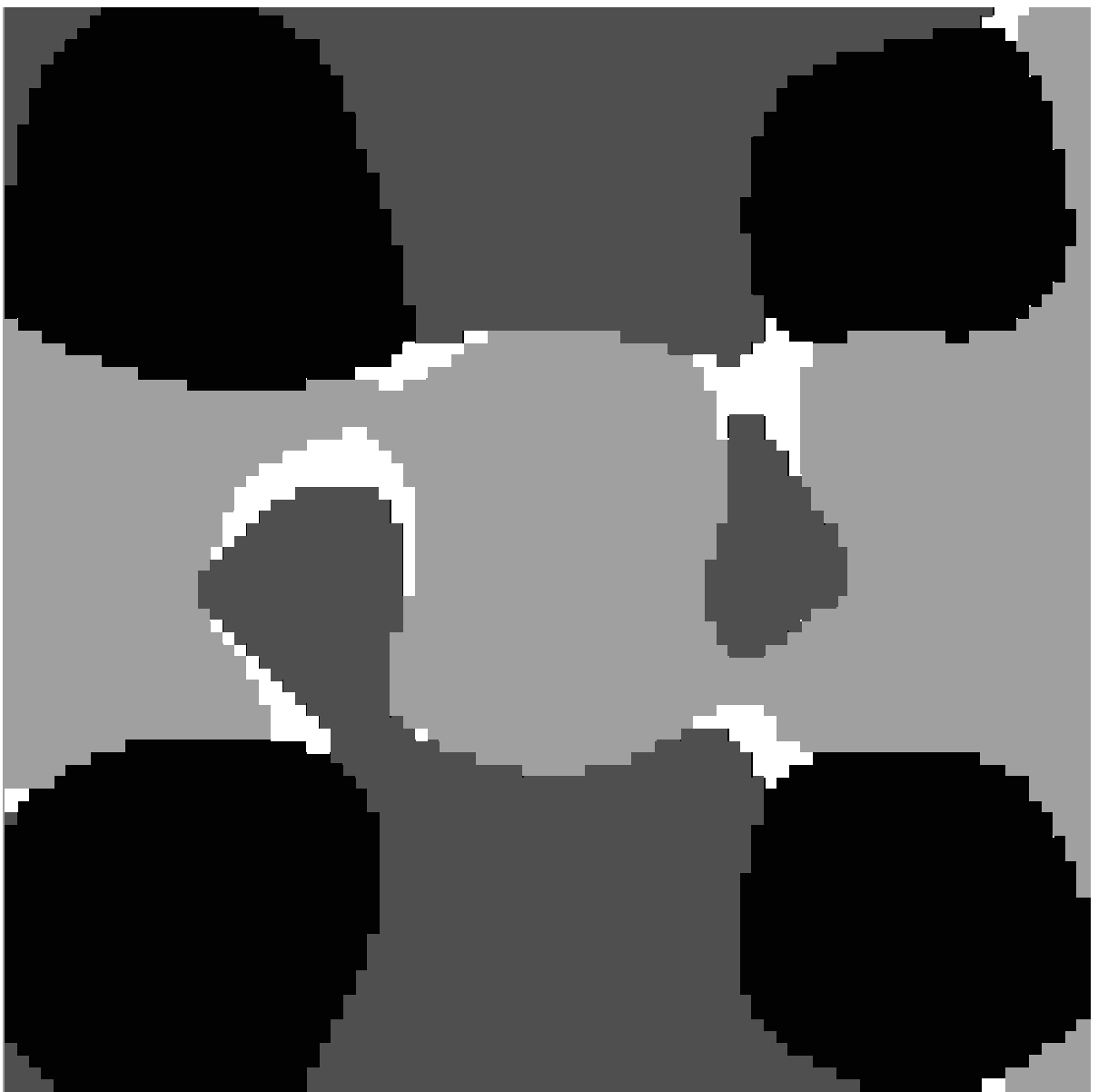}
		\end{minipage}%
	}%
	\subfigure[LSSMTC]{
		\begin{minipage}[t]{0.25\linewidth}
			\centering
			\includegraphics[width=1in]{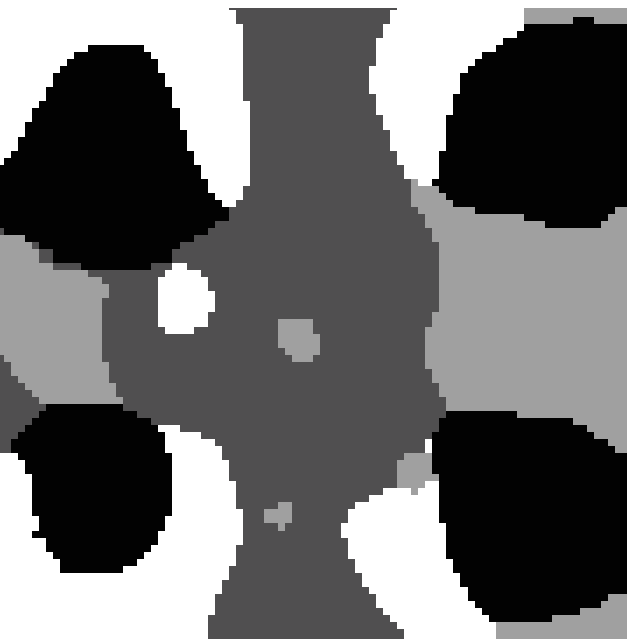}
		\end{minipage}
	}%
	\subfigure[CombKM]{
		\begin{minipage}[t]{0.25\linewidth}
			\centering
			\includegraphics[width=1in]{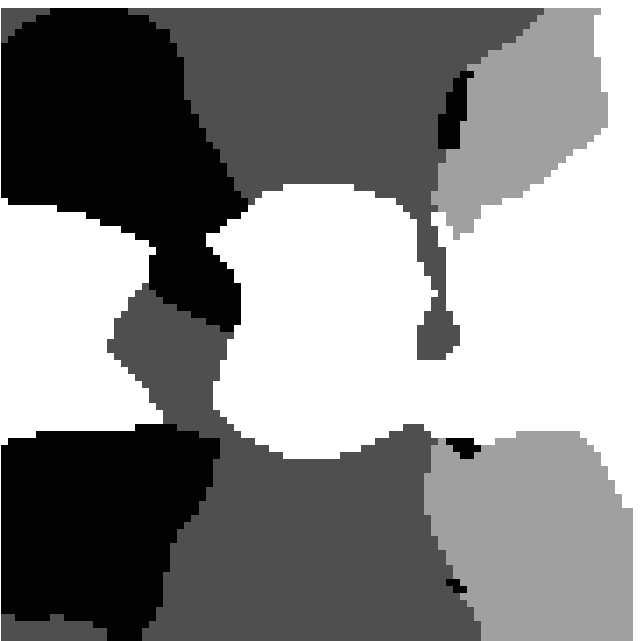}
		\end{minipage}
	}%
	\subfigure[TSC]{
		\begin{minipage}[t]{0.25\linewidth}
			\centering
			\includegraphics[width=1in]{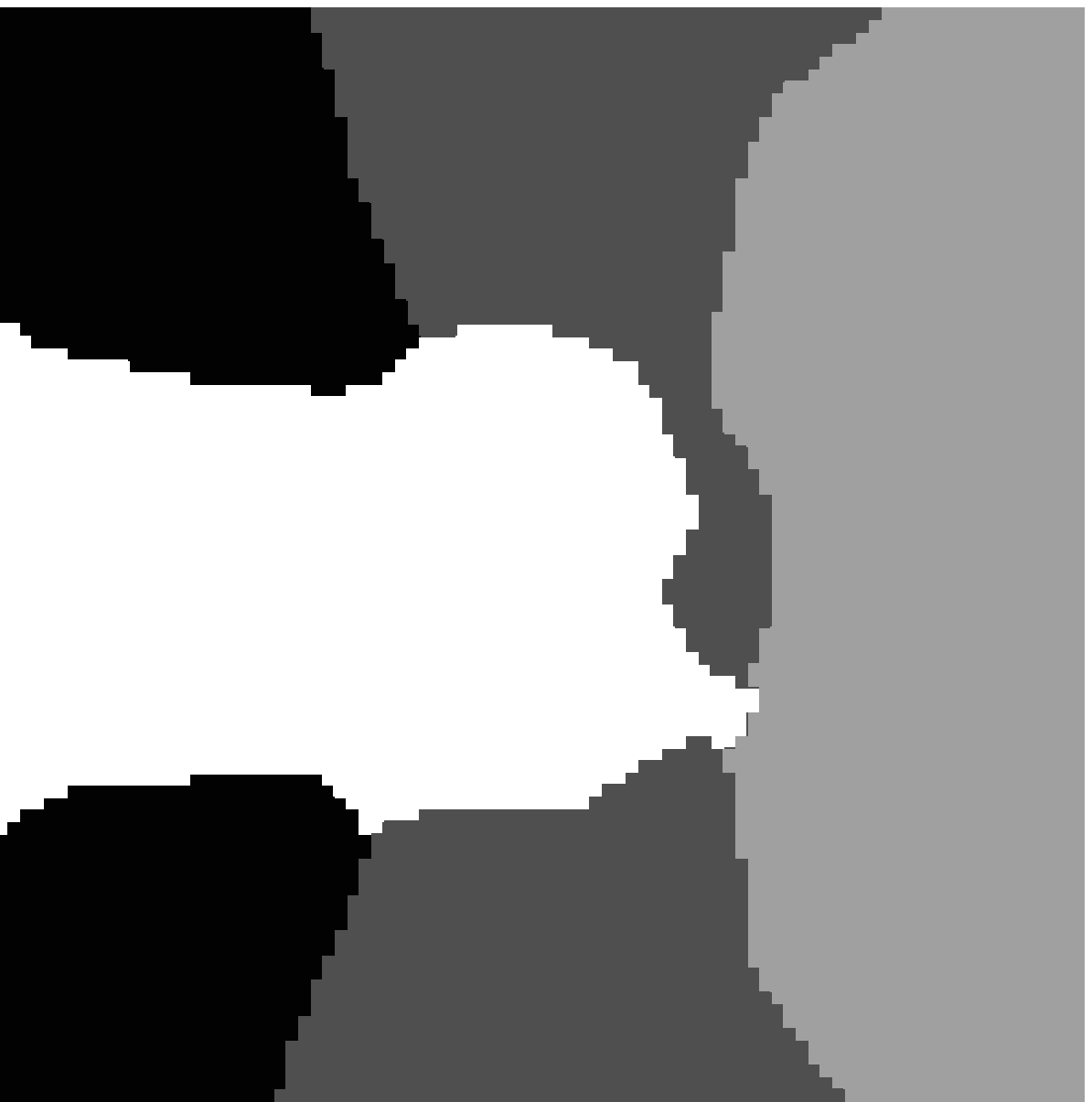}
		\end{minipage}
	}%
\quad
\subfigure[TDEM]{
	\begin{minipage}[t]{0.25\linewidth}
		\centering
		\includegraphics[width=1in]{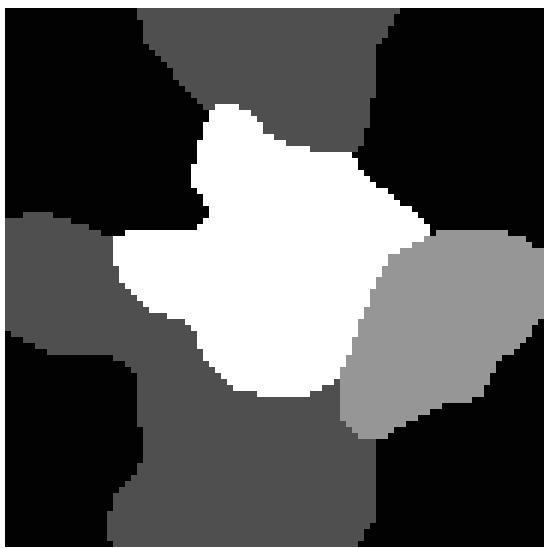}
	\end{minipage}
}%
	\subfigure[TFCM]{
		\begin{minipage}[t]{0.25\linewidth}
			\centering
			\includegraphics[width=1in]{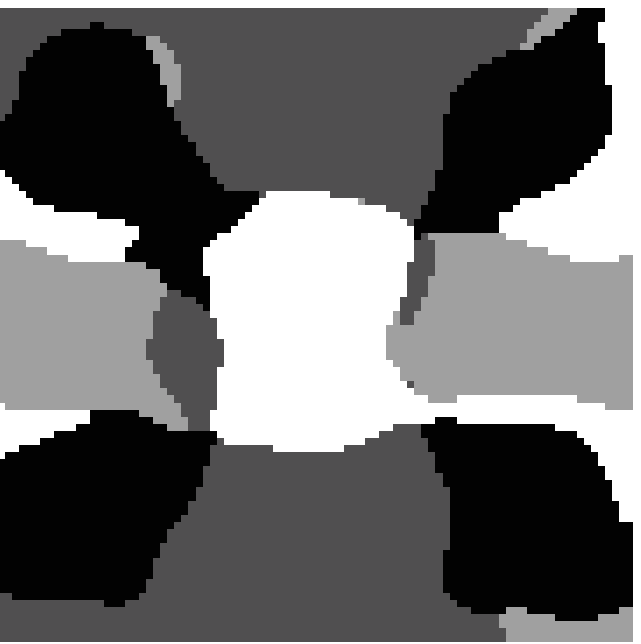}
		\end{minipage}
	}%
    \subfigure[TLEC]{
	\begin{minipage}[t]{0.25\linewidth}
		\centering
		\includegraphics[width=1in]{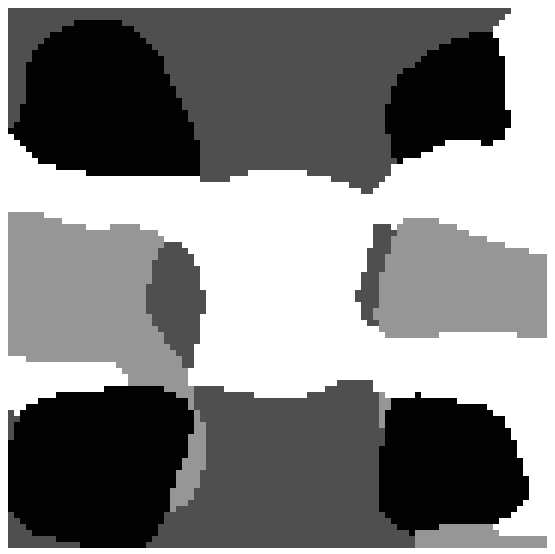}
	\end{minipage}
}%
	\subfigure[TECM]{
		\begin{minipage}[t]{0.25\linewidth}
			\centering
			\includegraphics[width=1in]{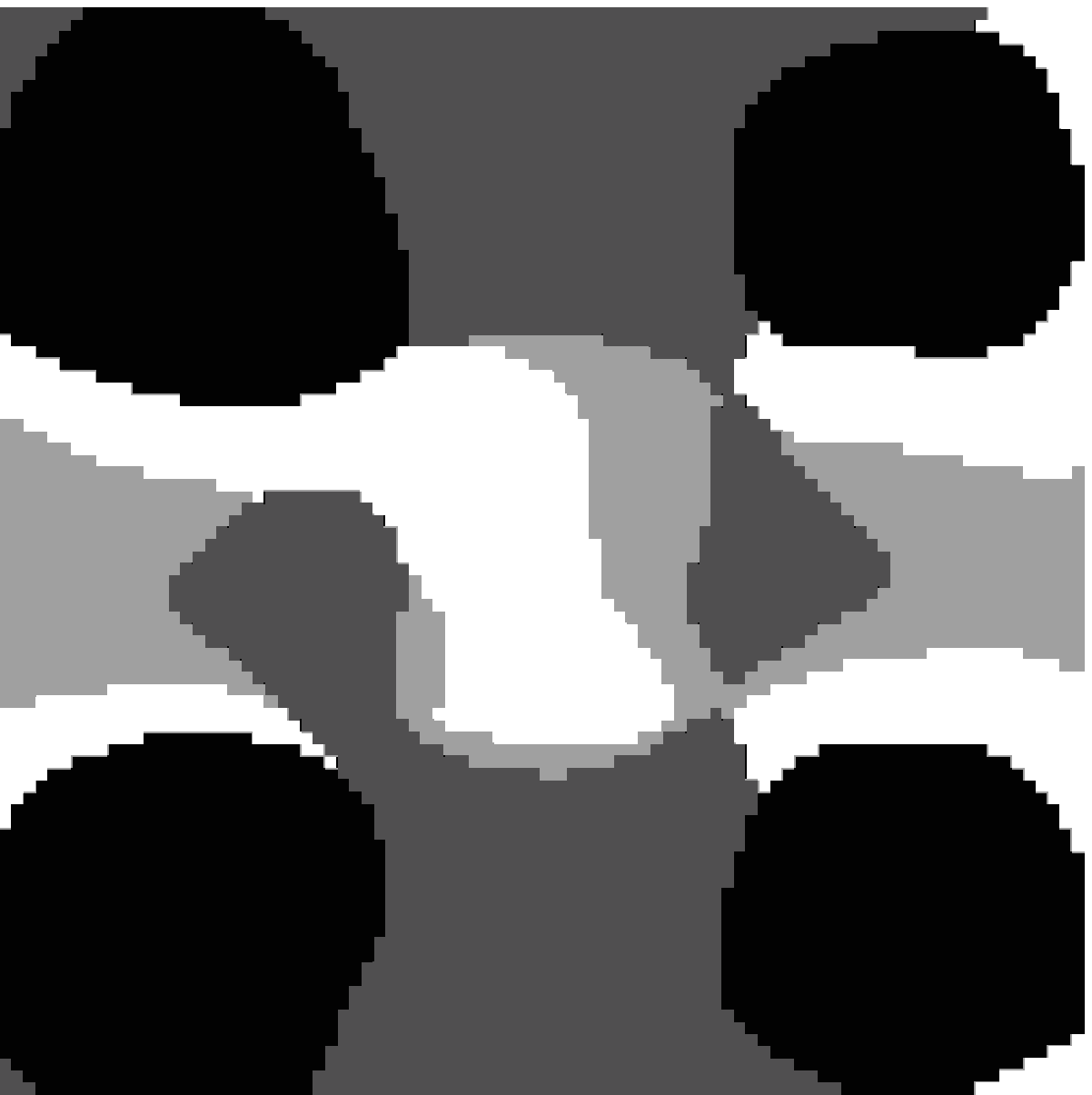}
		\end{minipage}
	}%

	\centering
	\caption{Segmentation results of T3-2 obtained by eight clustering algorithms.}
	\label{fig12}
\end{figure}

\begin{figure}[H]
	\centering
	\subfigure[ECM]{
		\begin{minipage}[t]{0.3\linewidth}
			\centering
			\includegraphics[width=0.98in]{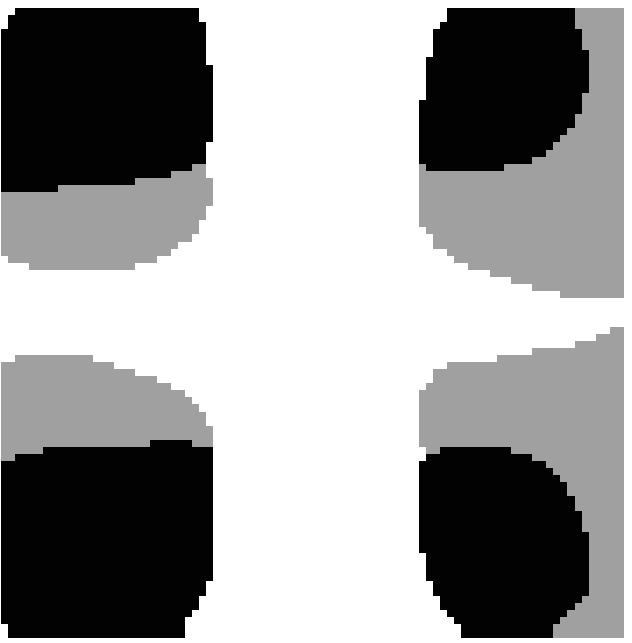}
		\end{minipage}%
	}%
	\subfigure[TFCM]{
		\begin{minipage}[t]{0.3\linewidth}
			\centering
			\includegraphics[width=1in]{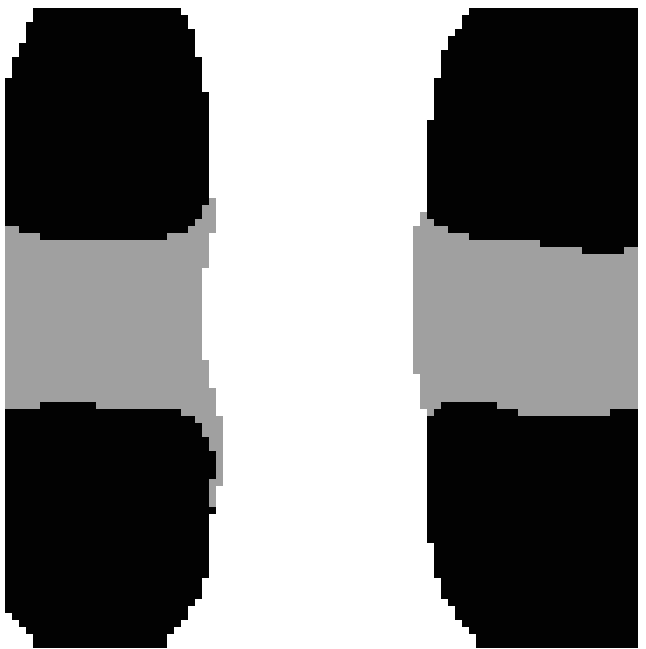}
		\end{minipage}
	}%
	\subfigure[TECM]{
		\begin{minipage}[t]{0.3\linewidth}
			\centering
			\includegraphics[width=1in]{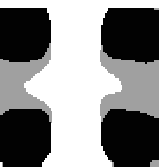}
		\end{minipage}
	}%
	\centering
	\caption{Segmentation results of T3-3 obtained by three clustering algorithms.}
	\label{fig13}
\end{figure}

\begin{figure}[H]
	\centering
	\subfigure[ECM]{
		\begin{minipage}[t]{0.3\linewidth}
			\centering
			\includegraphics[width=1in]{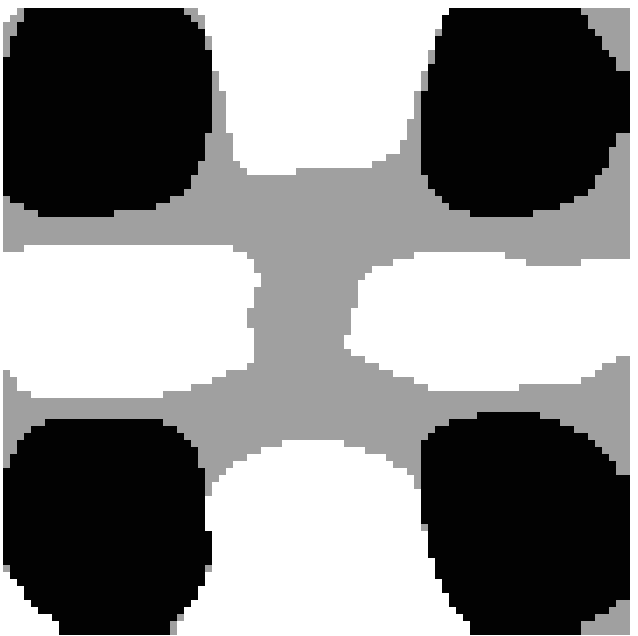}
		\end{minipage}%
	}%
	\subfigure[TFCM]{
		\begin{minipage}[t]{0.3\linewidth}
			\centering
			\includegraphics[width=1in]{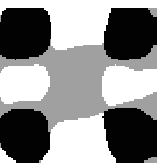}
		\end{minipage}
	}%
	\subfigure[TECM]{
		\begin{minipage}[t]{0.3\linewidth}
			\centering
			\includegraphics[width=1in]{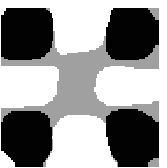}
		\end{minipage}
	}%
	\centering
	\caption{Segmentation results of T3-4 obtained by three clustering algorithms.}
	\label{fig14}
\end{figure}

\begin{figure}[H]
	\centering
	\subfigure[ECM]{
		\begin{minipage}[t]{0.3\linewidth}
			\centering
			\includegraphics[width=1.03in]{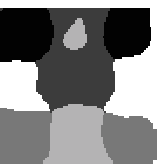}
		\end{minipage}%
	}%
	\subfigure[TFCM]{
		\begin{minipage}[t]{0.3\linewidth}
			\centering
			\includegraphics[width=1in]{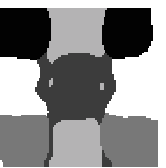}
		\end{minipage}
	}%
	\subfigure[TECM]{
		\begin{minipage}[t]{0.3\linewidth}
			\centering
			\includegraphics[width=1.01in]{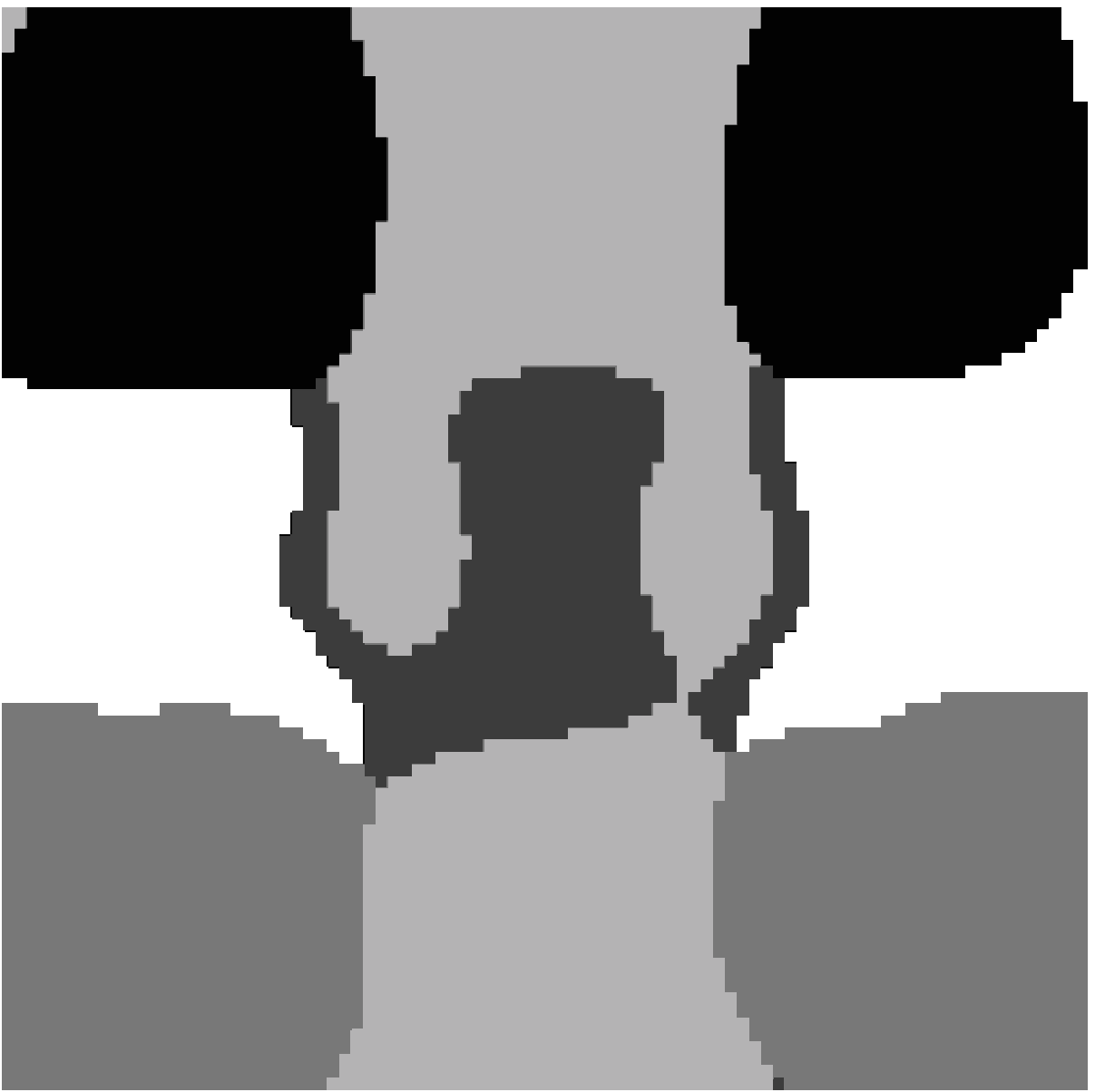}
		\end{minipage}
	}%
	\centering
	\caption{Segmentation results of T3-5 obtained by three clustering algorithms.}
	\label{fig15}
\end{figure}

\begin{figure}[H]
	\centering
	\subfigure[ECM]{
		\begin{minipage}[t]{0.3\linewidth}
			\centering
			\includegraphics[width=1.02in]{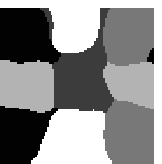}
		\end{minipage}%
	}%
	\subfigure[TFCM]{
		\begin{minipage}[t]{0.3\linewidth}
			\centering
			\includegraphics[width=1in]{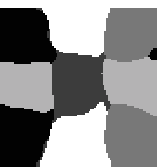}
		\end{minipage}
	}%
	\subfigure[TECM]{
		\begin{minipage}[t]{0.3\linewidth}
			\centering
			\includegraphics[width=1.02in]{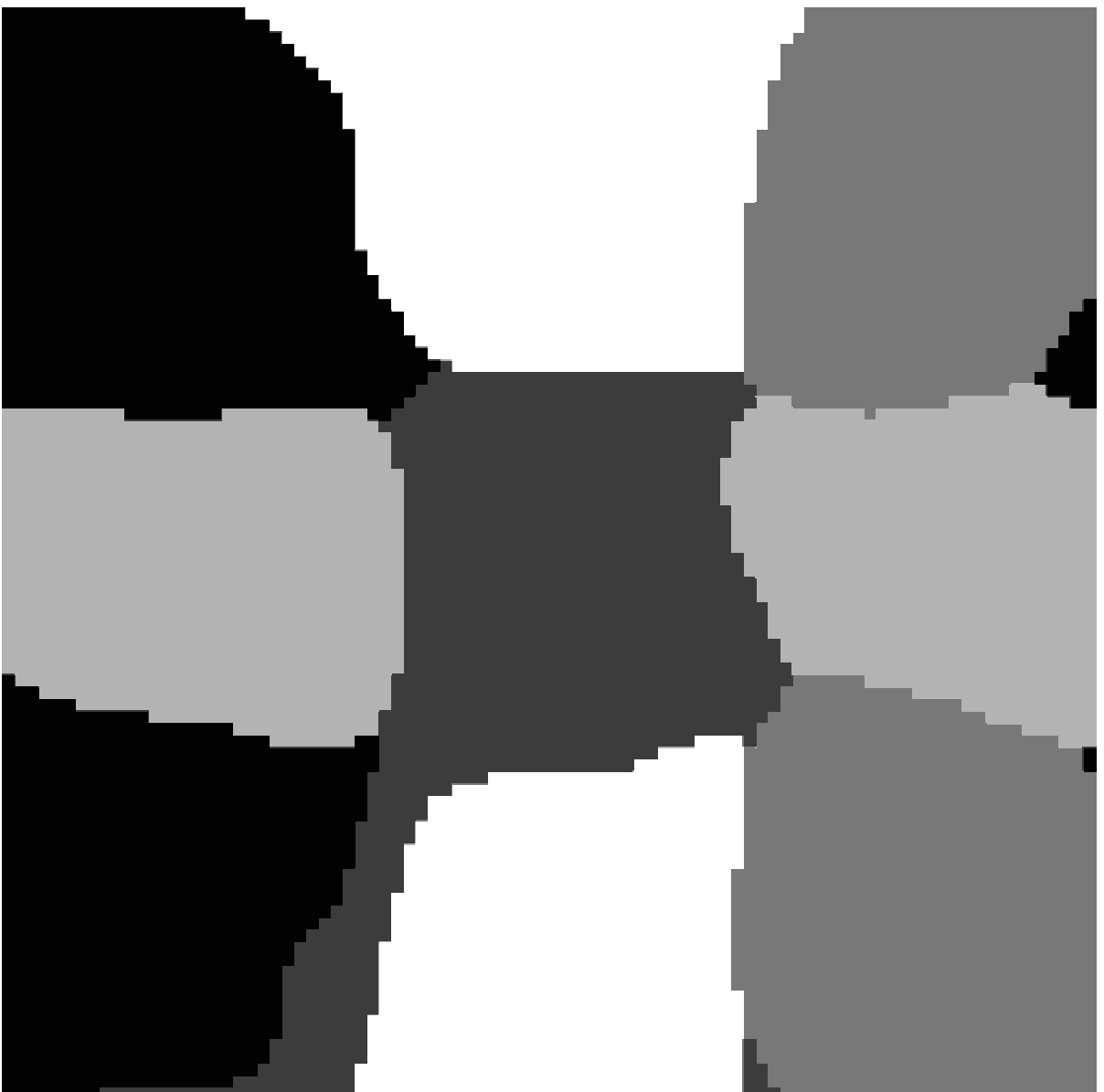}
		\end{minipage}
	}%
	\centering
	\caption{Segmentation results of T3-6 obtained by three clustering algorithms.}
	\label{fig16}
\end{figure}

Fig.~\ref{fig11}--Fig.~\ref{fig16} display the segmentation results for each target image using the different algorithms. In addition, Table~\ref{tab7} provides the quantitative scores of $ac$, $RI$, and $NMI$ obtained using the different algorithms. For the same reasons as those indicated in Section \ref{5.2}, the results of LSSMTC, CombKM, TSC, TEDM, and TLEC for certain datasets are absent. Based on the above results, the following conclusions can be drawn.

1) Owing to the existence of noise, ECM could not achieve acceptable texture segmentation performance on noisy target images. Conversely, benefiting from the reference information across domains, TECM could obtain a relatively superior performance compared to ECM. This demonstrates the effectiveness of transfer clustering.

2) TECM and TFCM could obtain relatively superior performance compared to the other multitask and transfer-clustering algorithms. Owing to the existence of noise and differences in data distribution between the two domains, raw source data were unsuitable for direct use in the clustering of the target data. Therefore, LSSMTC and CombKM could easily encounter negative effects unexpectedly between tasks, rather than simultaneously achieving performance improvements.

3) TECM and TFCM were similar in terms of the clustering indices of the hard partition. However, it can be observed from the experimental results that in the cases where TFCM was superior, the clustering performance gap between TFCM and TECM was small. However, in the cases where TECM was superior, the clustering performance gap between TECM and TFCM was large, which indicates that the proposed TECM was more powerful than TFCM overall. In addition, TECM had an advantage over TFCM in that it provided a deeper insight into the data when the data distribution was complex, as demonstrated in Section \ref{5.2.2}.

4) Regardless of whether the number of clusters in the source domain was equal to, greater than, or less than the number of clusters in the target domain, the TECM clustering performance always improved in comparison with ECM. Therefore, if the structure of the source data is partially similar to that of the target data, the learned knowledge from the source data will be beneficial for clustering the target data.

\begin{table}[H]
	\centering
	\caption{Clustering performance comparisons of six algorithms on T3-1 to T3-6}
	\resizebox{120mm}{50mm}{
		\begin{threeparttable}
			\begin{tabular}{|c|c|c|c|c|c|c|c|c|c|}
				\hline
				\multirow{2}[4]{*}{\textbf{Datasets}} & \multirow{2}[4]{*}{\textbf{Metrics}} & \multicolumn{8}{c|}{\textcolor[rgb]{ .2,  .2,  .2}{\textbf{ Algorithms}}} \bigstrut\\
				\cline{3-10}                 &              & ECM          & LSSMTC       & CombKM       & TSC    &TDEM     & TFCM         & TLEC &TECM \bigstrut\\
				\hline
				\multirow{3}[6]{*}{T3-1 (S3-1$\Rightarrow$T3-1)} & mean\_ac     & 0.705        & 0.664        & 0.646        & 0.690  &0.805      & 0.793        & 0.729 &\textbf{0.836} \bigstrut\\
				\cline{2-10}                 & mean\_RI     & 0.815        & 0.751        & 0.749        & 0.721   &0.822     & 0.818        & 0.764 &\textbf{0.865} \bigstrut\\
				\cline{2-10}                 & mean\_NMI    & 0.536        & 0.472        & 0.443        & 0.380   &0.586     & 0.579        & 0.480 &\textbf{0.629 } \bigstrut\\
				\hline
				\multirow{3}[6]{*}{T3-2 (S3-1$\Rightarrow$T3-2)} & mean\_ac     & 0.722        & 0.604        & 0.561        & 0.677   &0.718     & 0.735        & 0.685 &\textbf{0.742 } \bigstrut\\
				\cline{2-10}                 & mean\_RI     & 0.786        & 0.726        & 0.715        & 0.723    &0.770    & 0.794        & 0.733 &\textbf{0.828 } \bigstrut\\
				\cline{2-10}                 & mean\_NMI    & 0.463        & 0.401        & 0.368        & 0.406   &0.417     & 0.480        & 0.424 &\textbf{0.503 } \bigstrut\\
				\hline
				\multirow{3}[6]{*}{T3-3 (S3-1$\Rightarrow$T3-3)} & mean\_ac     & 0.774        & \textbackslash{} & \textbackslash{} & \textbackslash{}  & \textbackslash{}      & \textbf{0.918 } &\textbackslash{} &0.916 \bigstrut\\
				\cline{2-10}                 & mean\_RI     & 0.785        & \textbackslash{} & \textbackslash{} & \textbackslash{}  & \textbackslash{}      & 0.895        & \textbackslash{}&\textbf{0.901 } \bigstrut\\
				\cline{2-10}                 & mean\_NMI    & 0.550        & \textbackslash{} & \textbackslash{} & \textbackslash{}    & \textbackslash{}    & \textbf{0.734 } & \textbackslash{} &0.700 \bigstrut\\
				\hline
				\multirow{3}[6]{*}{T3-4 (S3-1$\Rightarrow$T3-4)} & mean\_ac     & 0.814        & \textbackslash{} & \textbackslash{} & \textbackslash{}  & \textbackslash{}     & 0.814        & \textbackslash{} &\textbf{0.840 }\bigstrut\\
				\cline{2-10}                 & mean\_RI     & 0.806        & \textbackslash{} & \textbackslash{} & \textbackslash{}     & \textbackslash{}   & 0.806        & \textbackslash{} &\textbf{0.830 }\bigstrut\\
				\cline{2-10}                 & mean\_NMI    & 0.566        & \textbackslash{} & \textbackslash{} & \textbackslash{}   & \textbackslash{}    & \textbf{0.566 } & \textbackslash{} &0.564 \bigstrut\\
				\hline
				\multirow{3}[6]{*}{T3-5 (S3-1$\Rightarrow$T3-5)} & mean\_ac     & 0.800        & \textbackslash{} & \textbackslash{} & \textbackslash{}& \textbackslash{} & \textbf{0.879 } & \textbackslash{} &0.877 \bigstrut\\
				\cline{2-10}                 & mean\_RI     & 0.875        & \textbackslash{} & \textbackslash{} & \textbackslash{} & \textbackslash{}& \textbf{0.916 } & \textbackslash{}&\textbf{0.916 } \bigstrut\\
				\cline{2-10}                 & mean\_NMI    & 0.671        & \textbackslash{} & \textbackslash{} & \textbackslash{}& \textbackslash{} & \textbf{0.755 } & \textbackslash{}  &0.738\bigstrut\\
				\hline
				\multirow{3}[6]{*}{T3-6 (S3-1$\Rightarrow$T3-6)} & mean\_ac     & 0.874        &\textbackslash{}              &\textbackslash{}              &\textbackslash{}   & \textbackslash{}           & \textbf{0.902 } & \textbackslash{} &0.890 \bigstrut\\
				\cline{2-10}                 & mean\_RI     & 0.910        &\textbackslash{}              & \textbackslash{}             &\textbackslash{}       & \textbackslash{}       & \textbf{0.925 } & \textbackslash{} &0.920 \bigstrut\\
				\cline{2-10}                 & mean\_NMI    & 0.707        &\textbackslash{}              &\textbackslash{}              & \textbackslash{}      & \textbackslash{}       & \textbf{0.751 } & \textbackslash{} &0.737 \bigstrut\\
				\hline
			\end{tabular}%
			\begin{tablenotes}
				\footnotesize
				\item[*] For a $\Rightarrow$ b, a and b denote the source and target datasets, respectively.
			\end{tablenotes}
		\end{threeparttable}
	}
	\label{tab7}%
\end{table}%

\subsection{Parameter analysis} \label{5.4}
The balance coefficient of transfer learning $\lambda$ is an important parameter in the TECM algorithm, which significantly influences the algorithm performance. In this section, the influence of $\lambda$ on the TECM clustering performance is discussed. Here, the synthesis datasets described in Table~\ref{three} were used to study the variation in the algorithm performance with $\lambda$. The performance of the algorithm was evaluated using $ac$.

\begin{figure}[H]
	\centering
	\begin{threeparttable}	
		\includegraphics[width=\linewidth]{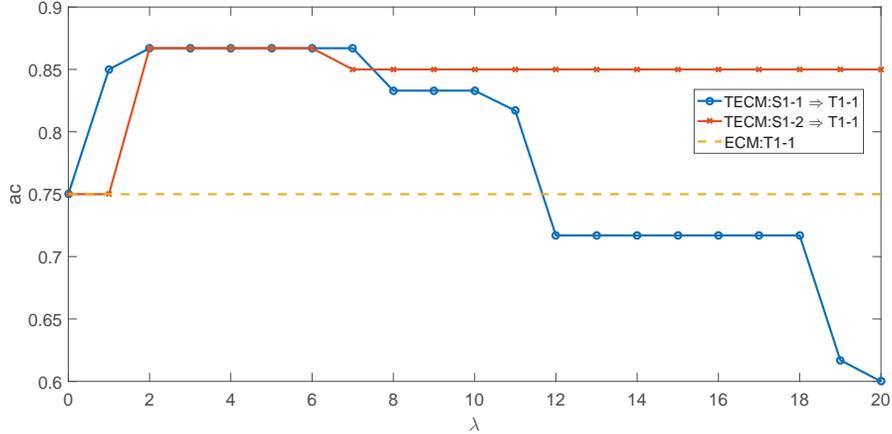}
	\end{threeparttable}
	\caption{TECM clustering performance with different values of $\lambda$.}
	\label{fig17}
\end{figure}

Fig.~\ref{fig17} indicates that the TECM clustering performance first improves and then decreases with an increase in $\lambda$. When the value of $\lambda$ is relatively small, the clustering performance is not significantly improved compared to ECM because the transferred knowledge has a small role. When the value of $\lambda$ is relatively large, the clustering performance is even worse than that of ECM because the clustering is completely dependent on the knowledge learned from the source data, which results in a negative transfer. In practice, the grid search strategy can be used to determine an effective value of $\lambda$ by optimizing an internal clustering index.

\section{Conclusions} \label{six}
In this study, a TECM clustering algorithm was proposed to improve the clustering performance of the target data with the help of barycenters learned from the source data for situations where target data are insufficient or contaminated. The proposed algorithm is applicable to situations where the source and target domains have the same or different numbers of clusters. In synthetic and real dataset experiments, the proposed TECM achieved approximately 5$\%$ and 7$\%$ performance improvement over ECM for insufficient and contaminated target data on average, respectively. In addition, compared with other related multitask or transfer-clustering algorithms, the proposed TECM performed the best overall.

In the future, this research topic can be further explored by integrating TECM with deep learning such that the clustering algorithm can be better applied to raw data such as images and texts.

\section*{Acknowledgments}
This work was funded by the National Natural Science Foundation of China (Grant Nos. 62171386, 61801386, and 61790552), Key Research and Development Program in Shaanxi Province of China (Grant No. 2022GY-081), and China Postdoctoral Science Foundation (Grant Nos. 2019M653743 and 2020T130538).

\section*{References}






\end{document}